\documentclass[twoside]{article}

\usepackage[accepted]{aistats2025}
\usepackage[round]{natbib}
\renewcommand{\bibname}{References}
\renewcommand{\bibsection}{\subsubsection*{\bibname}}

\usepackage{amsmath,amsfonts,bm,amsthm}
\usepackage{thmtools,thm-restate}

\newtheorem{thm}{Theorem}

\def\eqref#1{equation~\ref{#1}}

\def\1{\bm{1}}

\DeclareMathAlphabet{\mathsfit}{\encodingdefault}{\sfdefault}{m}{sl}
\SetMathAlphabet{\mathsfit}{bold}{\encodingdefault}{\sfdefault}{bx}{n}

\def\gO{{\mathcal{O}}}

\usepackage[T1]{fontenc}
\usepackage{lmodern}

\usepackage{booktabs}       
\usepackage{graphicx} 
\usepackage{subcaption}
\usepackage{algpseudocode}
\usepackage{algorithm}
\usepackage{float}
\usepackage{caption}
\usepackage{amsmath}
\usepackage{amssymb}
\usepackage{hyperref}
\usepackage{cleveref} 
\usepackage{multirow}
\usepackage{url}

\newcommand{\av}{\operatorname{av}}

\begin{document}

%

%

\runningtitle{Efficient Trajectory Inference in Wasserstein Space Using Consecutive Averaging}
\runningauthor{Amartya Banerjee, Harlin Lee, Nir Sharon, Caroline Moosmüller}

\twocolumn[

\aistatstitle{Efficient Trajectory Inference in Wasserstein Space Using Consecutive Averaging}

\aistatsauthor{%
Amartya Banerjee \And
Harlin Lee \And 
Nir Sharon \And
Caroline Moosmüller}

\aistatsaddress{%
\footnotesize Dept. of Computer Science\\ UNC Chapel Hill \And
\footnotesize SDSS\\ UNC Chapel Hill \And 
\footnotesize Dept. of Applied Math\\ Tel Aviv University \And
\footnotesize Dept. of Mathematics\\ UNC Chapel Hill
}

]

\begin{abstract}
    Capturing data from dynamic processes through cross-sectional measurements is seen in many fields, such as computational biology. Trajectory inference deals with the challenge of reconstructing continuous processes from such observations. In this work, we propose methods for B-spline approximation and interpolation of point clouds through consecutive averaging that is intrinsic to the Wasserstein space. Combining subdivision schemes with optimal transport-based geodesic, our methods carry out trajectory inference at a chosen level of precision and smoothness, and can automatically handle scenarios where particles undergo division over time. We prove linear convergence rates and rigorously evaluate our method on cell data characterized by bifurcations, merges, and trajectory splitting scenarios like \textit{supercells}, comparing its performance against state-of-the-art trajectory inference and interpolation methods. The results not only underscore the effectiveness of our method in inferring trajectories but also highlight the benefit of performing interpolation and approximation that respect the inherent geometric properties of the data.
\end{abstract}

\section{INTRODUCTION}

Many dynamic processes yield cross-sectional observations at different time steps, which can be represented as a sequence of \textit{point clouds}, or discrete probability measures. To understand the underlying process, we need to first interpolate or approximate a likely path between these measurements. This problem of reconstructing continuous paths is called \textit{trajectory inference}, which is seen in applications such as computational biology (\cite{sha2023reconstructing,tong2020trajectorynet,huguet2022manifold,schiebinger2019optimal,saelens2019comparison}), computer graphics~(\cite{huang2022representation}), and control theory~(\cite{howe2022myriad,craig2024blob}). 
This paper considers the case where we only observe the state of a population as a whole, i.e. points in a point cloud do not have an inherent order. It may be the case that each sample is equivalent and interchangeable (e.g. drones in a swarm), or that there is no preservation of samples across time (e.g. cell measurement). 

Recently, \textit{optimal transport (OT)} (\cite{peyre2019computational,ambrosio2013user,villani2008optimal}) has shown to be effective in this problem, as it can provide a matching between two point clouds. By using OT to infer which sample at time step $t_i$ could correspond to which one at $t_j$, trajectories can be constructed in a way that minimize cost or effort. 
In the seminal work by \cite{schiebinger2019optimal}, a piecewise linear OT interpolation method is proposed to infer cell trajectories. Higher order piecewise polynomials (e.g. cubic splines) were introduced in \cite{chen2018measure,benamou2019cubicsplines} and extended by \cite{chewi2021fast,clancy2022wassersteinfisherrao,justiniano2023approximation}. 
Deep learning-based methods that try to fit smoother paths include TrajectoryNet (\cite{tong2020trajectorynet}) and MIOFlow (\cite{huguet2022manifold}), which train neural ODEs with OT-informed tools, and CFM (\cite{CFM_tong2023improving}), which uses flow-based methods for constructing trajectories. In a similar spirit, WLF (\cite{wlflow_neklyudov_computational_2023}) proposes a unified approach for optimal transport based trajectory inference by optimizing a Lagrangian action functional in the space of densities, allowing one to incorporate problem-specific priors directly into the learned dynamics.

Our paper fits in this line of work in the following sense. We propose a set of algorithms for trajectory inference (either interpolatory or approximate) on point clouds  using consecutive OT-based geodesic averaging in the Wasserstein space. The methods are flexible and efficient as they allow us to tune the trajectories' smoothness and accuracy. Furthermore, they are \textit{intrinsic to the geometry of the Wasserstein space}, which naturally handles mass-splitting phenomena. 

We first propose the \emph{Wasserstein Lane-Riesenfeld (WLR)} algorithm to approximate B-splines in the Wasserstein space. We leverage the algorithm by \cite{Lane1980LRalgorithm}, which efficiently defines B-splines in $\mathbb{R}^d$ through consecutive averaging. It has been shown that this algorithm can be extended to Riemannian manifolds in \cite{wallner2005subdivision}, and our method extends this idea to the Wasserstein space by using a OT-based geodesic.

While the WLR algorithm is the main method we propose, we observe that consecutive averaging can be applied more broadly beyond B-splines, and following ideas from \cite{wallner2005subdivision}, we showcase an interpolatory trajectory inference method based on the 4-point scheme by \cite{DYN1987pointscheme}. We further discuss how our idea extends to any iterative method that uses consecutive averaging (e.g.\ any subdivision scheme).

\begin{figure}[htp]
    \centering
    \includegraphics[width=\linewidth]{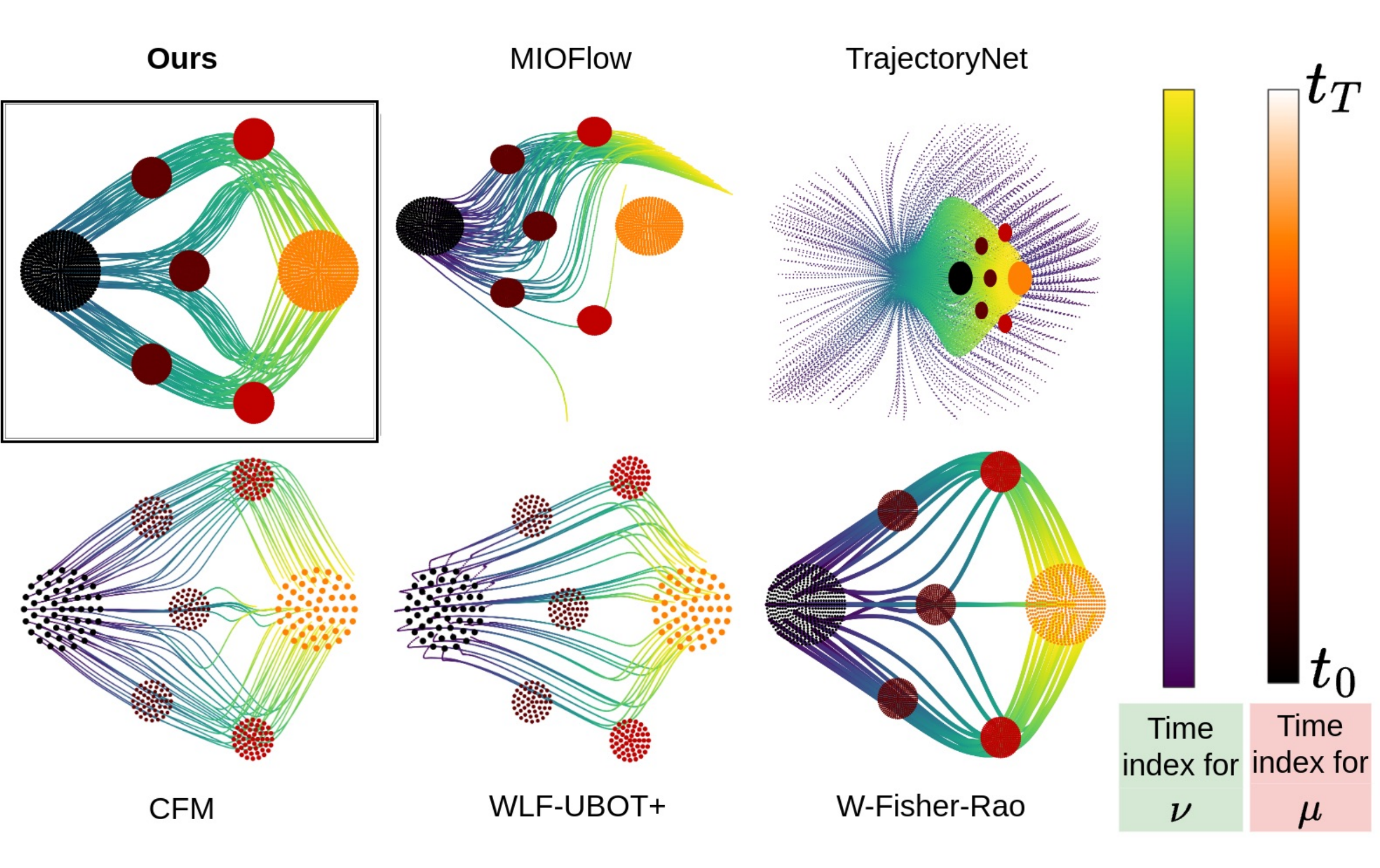}
    \caption{Our proposed WLR successfully performs trajectory inference on Converging Gaussian dataset, which has different number of points per time step. While others maintain a fixed number of trajectories from initialization to termination, our method can split trajectories automatically.
    The fact that WLR can naturally deal with mass splitting phenomena is one of its major benefits. See Sec. \ref{sec:experiments} for details.}
    \label{fig:failure_cases}
\end{figure}

Our contributions are:
\begin{enumerate}
    \item Algorithmically, 
   we define curves intrinsic to the Wasserstein geometry that do not depend on splines in $\mathbb{R}^d$ 
   as required in \cite{chewi2021fast}. Unlike \cite{chewi2021fast}, which only works on uniform data in $\mathbb{R}^n$ by applying cubic splines after a fixed OT matching, our WLR approach handles non-uniform or unbalanced masses and easily adapts to arbitrary support spaces. Moreover, while methods like \cite{benamou2019cubicsplines,chen2018measure} formulate a global cubic spline minimization in Wasserstein space (with existence results largely demonstrated on Gaussian examples), solving such an optimization is not tractable in general. In contrast, WLR is a simple iterative scheme that remains fully intrinsic to Wasserstein geometry but is still straightforward to implement for real data. Our algorithms can therefore naturally deal with ``trajectory splitting'', which is needed when mass is non-uniformly distributed over point clouds. This makes our methods comparable to other intrinsic Wasserstein spline methods, such as \cite{chen2018measure,benamou2019cubicsplines}. Furthermore, our proposed algorithms can flexibly handle both approximation (via B-splines) and exact interpolation (via the 4-point scheme), and can be easily extended to other subdivision schemes. 
    \item Theoretically, our approach is consistent with existing theory in degenerate cases. When each point cloud consists of just one point, we recover classical B-spline theory (on $\mathbb{R}^d$). When there are only two point clouds, we recover the OT assignment between those point clouds. Importantly, unlike prior spline-subdivision work on finite-dimensional Riemannian manifolds (e.g.,\cite{wallner2005subdivision} and \cite{sharon2017global}), we prove convergence for an iterative subdivision scheme in the \textit{infinite-dimensional} Wasserstein space.
     \item Experimentally, WLR can accurately infer individual trajectories between uniform point clouds as for example needed in cell trajectories. Moreover, it can handle complex scenarios involving weighted mass splitting as in the case of \textit{supercell} data. 
    Subject to user's choice of parameters, WLR is fast and stable compared to neural ODE-based methods such as \cite{tong2020trajectorynet,huguet2022manifold}, which can fail under stiff dynamics or unstable optimization. 
     WLR is however slightly slower than extrinsic methods such as \cite{chewi2021fast}, as we need to solve more OT problems. 
\end{enumerate}

We also mention the concurrent publication \cite{baccou2024subdivision}, which, similar to this paper, combines subdivision schemes with optimal transport. We present a more comprehensive comparison to other trajectory inference methods and provide new applications to cell data, including supercells.

In the following sections, we describe background on optimal transport and splines (Sec. \ref{sec:background}), state (Sec. \ref{sec:method}) then demonstrate (Sec. \ref{sec:experiments}) our proposed algorithms on various numerical experiments.


\section{PRELIMINARIES} \label{sec:background}
We define relevant notations and formulate the trajectory inference problem, then present background on optimal transport and splines in the context of that problem. We view optimal transport as a special case of trajectory inference where there are only two point clouds, and splines in $\mathbb{R}^d$ as another edge case where there is only one point per point cloud.

\subsection{Problem formulation} \label{sec:formulation}

We consider probability measures $\mu_t$ over $\mathbb{R}^d$ depending on a continuous time parameter $t$. For the purpose of implementation, $\mu_t$ is a discrete probability measure or \textit{point cloud}, i.e.\ it is defined by a set of support points $ x_t = \{x_{t,i}\}_{i=1}^{n_t} \subset \mathbb{R}^d$, and a probability vector $a_t \in \mathbb{R}^{n_t}_+$ with $\sum_{i=1}^{n_t}a_{t,i}=1$. Mathematically, this can be described as $\mu_t = \sum_{i=1}^{n_t}a_{t,i}\delta_{x_{t,i}}$; see \cite{peyre2019computational}. Here, $a_{t,i}$ describes the amount of mass located at $x_{t,i}$, and we allow for non-uniformly distributed mass. Note that the number of points $n_t$ may change at different times $t$, and while the points in $x_t$ are inherently unordered, we give them an arbitrary yet consistent ordering and represent them as a matrix $x_t \in \mathbb{R}^{d \times n_t}$ in a slight abuse of notation.

Given $T+1$ observations $\mu_{t_0},\ldots,\mu_{t_T}$, where $t_0< \ldots <t_T$, the aim of \textit{trajectory inference} is to find probability measures $\nu_t$ such that for $j=0,\ldots, T$, $\nu_{t_j}\approx \mu_{t_j}$ for approximation, or such that $\nu_{t_j} = \mu_{t_j}$ for interpolation.

\subsection{Discrete optimal transport and the Wasserstein space}\label{sec:OT}

In this section, we consider a single pair of point clouds $(\mu_{t_0}, \mu_{t_1})$ as defined in Section \ref{sec:formulation}. Discrete optimal transport (OT) studies the problem of transporting $\mu_{t_0}$ to $\mu_{t_1}$ in an ``optimal'' manner. Given a cost matrix $C=[C_{ij}] \in \mathbb{R}_{+}^{n_{t_0}\times n_{t_1}}$ that describes how much effort is needed to move mass from one point in $\mu_{t_0}$ to another in $\mu_{t_1}$, OT aims to find a coupling matrix $\gamma \in \mathbb{R}_{+}^{n_{t_0}\times n_{t_1}}$ that attains
\begin{equation}
  f:=  \min_\gamma ~\langle \gamma, C \rangle, \;  \text{s.t.}~\gamma\mathbf{1} = a_{t_0},\gamma^\top\mathbf{1}  = a_{t_1},\gamma \ge 0.  \label{eq:kantorovich}
\end{equation}
Sometimes called the Kantorovich formulation, this equation  by \cite{Kantorovich2006_OptimalTransport} is often solved via linear programming. This has been crucial in applications including economics, inverse problems, image processing, computational biology, and most recently, machine learning (e.g. \cite{courty2017joint}); see \cite{peyre2019computational} for an excellent overview on computational OT.  
A popular choice for cost is $C_{ij}=\|x_{t_0, i}-x_{t_1, j}\|^p$ for $p\ge 1$, in which case $W_p(\mu_{t_0}, \mu_{t_1})$, the $p$-Wasserstein distance between $\mu_{t_0}$ and $\mu_{t_1}$, is defined as $f^{1/p}$ using the minimum value attained in \cref{eq:kantorovich}. 

When the space of probability measures on $\mathbb{R}^d$ with finite $p$-th moment is equipped with $W_p$, we call it \textit{Wasserstein space}. The Wasserstein space is a metric space and carries Riemannian-like structure. It can formally be considered an infinite dimensional Riemannian manifold (\cite{OttoRiemannianOT2001,LottWassersteinRiemannian2008}). 
Representing the point clouds to be on the Wasserstein space allows us to define useful geometric properties such as barycenters (\cite{agueh2011barycenters}). Of particular interest to trajectory inference is the \textit{geodesic} in the Wasserstein space.
Once $\gamma$ is obtained, the constant speed geodesic between $\mu_{t_0}$ and $\mu_{t_1}$ is given by the pushforward measure:
\begin{align} \label{eq:OT-geodesic}
   &\mu_\alpha = (\pi_\alpha)_{\sharp}\gamma, \quad \text{where } \\
   &\pi_\alpha(x_{t_0, i}, x_{t_1, j}) = (1-\alpha)x_{t_0, i}+\alpha \, x_{t_1, j} \text{ and }\alpha \in [0, 1].\nonumber
\end{align}
 $\mu_\alpha$ is the \underline{shortest} path between $\mu_{t_0}$ and $\mu_{t_1}$ in the Wasserstein space. See \cite{villani2008optimal,santambrogio2015optimal,ambrosio2013user} for more background on Wasserstein geometry.

\subsection{Splines through consecutive averaging}\label{sec:splines-through-averaging}

\begin{figure}[htp]
    \centering
    \begin{subfigure}[b]{0.24\linewidth}
        \includegraphics[trim=270 90 130 90, clip, width=\linewidth]{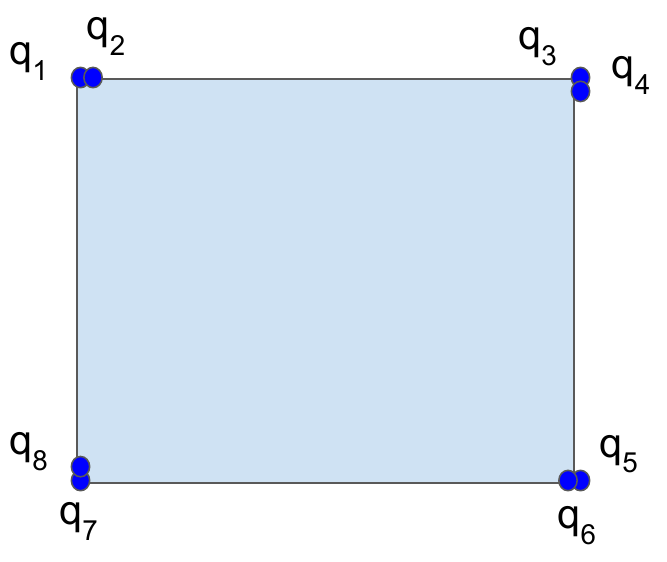}
        \caption{}
        \label{fig:lr2}
    \end{subfigure}%
    \begin{subfigure}[b]{0.24\linewidth}
        \includegraphics[trim=270 90 130 90, clip, width=\linewidth]{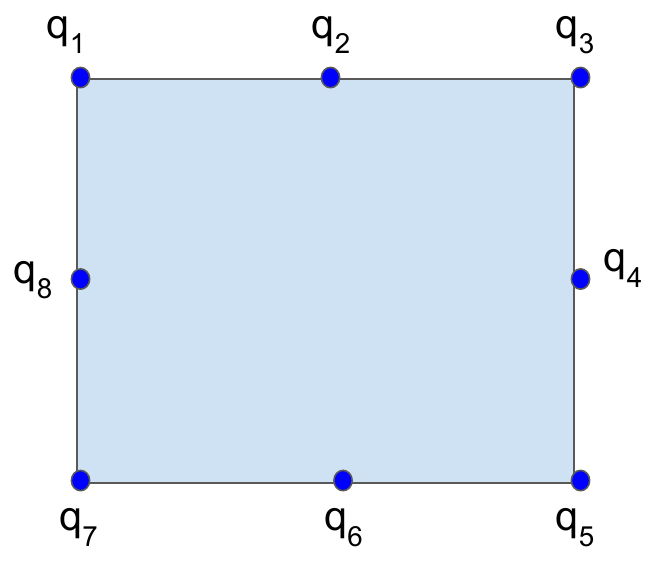}
        \caption{}
        \label{fig:lr3}
    \end{subfigure}%
    \begin{subfigure}[b]{0.24\linewidth}
        \includegraphics[trim=270 90 130 90, clip, width=\linewidth]{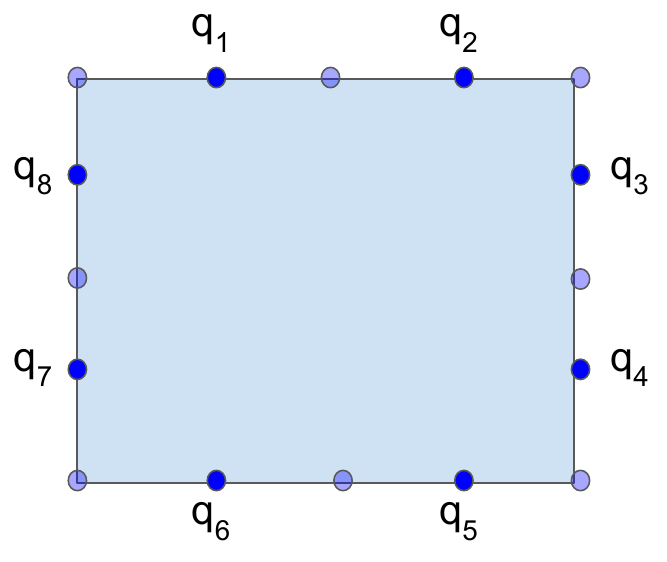}
        \caption{}
        \label{fig:lr4}
    \end{subfigure}%
    \begin{subfigure}[b]{0.24\linewidth}
        \includegraphics[trim=270 90 130 90, clip, width=\linewidth]{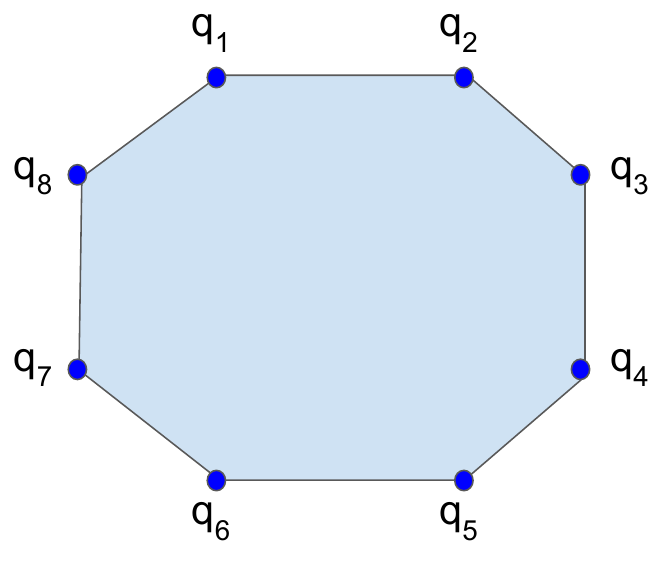}
        \caption{}
        \label{fig:lr6}
    \end{subfigure}
    \caption{Illustration of the classical Lane-Riesenfeld (Alg.~\ref{alg:LRm}) with $M=2$, $R=1$. This refinement step can be repeated to approximate a smooth cubic B-spline. (a) Doubling points; (b) First averaging; (c) Second averaging; (d) Refined points.}
    \label{fig:lane-riesenfeld}
\end{figure}

We discuss another edge case of Section \ref{sec:formulation}, namely where each point cloud only has a single point, that is, $n_t = 1$. Splines are popular methods that fit piecewise polynomial curves through such data points in $\mathbb{R}^d$. In this work, we limit our discussion of splines from the viewpoint of \textit{subdivision schemes}. These are iterative refinement methods that approximate continuous or smooth curves based on a sequence of initial data points, which is also called knots. See \Cref{app:subdivision-schemes} for details.

We focus on one such method by \cite{Lane1980LRalgorithm} in this work (Appendix \ref{app:subdivision-schemes} \Cref{alg:LRm}). The Lane-Riesenfeld algorithm depends on two parameters--- the degree $M$, which is related to the smoothness of the limit curve, and the refinement level $R$, which flexibly handles the accuracy of the approximation curve. As $R\to \infty$, we obtain a B-spline of degree $M$ in $\mathbb{R}^d$ which is $C^{M-1}$ (\cite{cavaretta1989stationary}).

The algorithm contains two main phases: doubling of points and computing averages between consecutive points. 
An illustration is provided in \Cref{fig:lane-riesenfeld} with degree $M=2$ and refinement level $R=1$. The algorithm is applied to a closed curve with four initial points, refining the shape iteratively.

Following \cite{wallner2005subdivision,sharon2017geodesic}, we can generalize the averaging phase via a linear averaging operator between two points $y_i,y_j \in \mathbb{R}^d$ and $\alpha \in [0,1]$:
\begin{equation}\label{eq:linear-average}
    \av_{\alpha}(y_i, y_j)= \left( 1-\alpha \right)y_i+\alpha\, y_j.
\end{equation}
With this operator, step (13) of  \Cref{alg:LRm} becomes $y_{j}^{(m)} \gets \av_{1/2}(y_{j}^{(m-1)},y_{j+1}^{(m-1)})$. It has been observed that the Lane-Riesenfeld algorithm can be further adapted to work on finite-dimensional Riemannian manifolds by replacing $\av_{\alpha}$ with the geodesic average on the Riemannian manifold (\cite{wallner2005subdivision}).

\section{WASSERSTEIN CURVE APPROXIMATION THROUGH CONSECUTIVE AVERAGING} \label{sec:method}

\begin{algorithm}[tp]
\caption{Wasserstein Lane-Riesenfeld}
\label{alg:WLRm}
\begin{algorithmic}[1]
\Procedure{WLR}{$[\mu_{t_j}]_{j=0}^T, R, M$}
\State \textbf{input} Point \textit{clouds} to be refined $ [\mu_{t_j}]_{j=0}^T$
\State \textbf{input} Refinement Level $R \in \mathbb{Z}_{+}$
\State \textbf{input} B-Spline degree $M$
\State $\nu^{(M)} \gets [\mu_{t_j}]_{j=0}^T$ 
\State \Comment{\textit{Initialize point clouds to be doubled}}
\For{$r =1$ \textbf{to} $R$ }
    \For{$j = 0$ \textbf{to} $|\nu^{(M)}|$} 
        \State $\nu_{2j}^{(0)} \gets \nu_j^{(M)}$        \Comment{\textit{Doubling point clouds}}
        \State $\nu_{2j+1}^{(0)} \gets \nu_j^{(M)}$
    \EndFor
    \State Duplicate $\nu_{0}^{(0)}$ and $\nu_{T}^{(0)}$ $m$ times.
    \For{$m = 1$ \textbf{to} $M$} 
        \For{$j = 0$ \textbf{to} $|\nu^{(m-1)}|$}
            \State $\nu_{j}^{(m)}  \gets \text{OT-av}(\nu_{j}^{(m-1)} , \nu_{j+1}^{(m-1)}, \frac{1}{2} )$ 
            \State \Comment{\textit{Repeated OT averaging}}
        \EndFor
    \EndFor
\EndFor
\State \Return Refined point clouds $\nu^{(M)}$ \EndProcedure
\end{algorithmic}
\end{algorithm}

We have now gathered two ingredients for trajectory inference. On one hand, we have a way to calculate the shortest path between two point clouds in the Wasserstein space (\Cref{sec:OT}). On the other hand, we have a way to efficiently fit a curve through many points  (not point \textit{clouds}) in $\mathbb{R}^d$ from \Cref{sec:splines-through-averaging}. Our key insight is that if we replace the linear averaging operator in \cref{eq:linear-average} with the Wasserstein geodesic in \cref{eq:OT-geodesic}, we can use the consecutive averaging approach described in \Cref{sec:splines-through-averaging} to define smooth approximating curves through the point clouds in the Wasserstein space.

We propose the Wasserstein Lane-Riesenfeld (WLR) algorithm (c.f. \Cref{alg:WLRm}), which parallels the ideas of \cite{sharon2017global,sharon2017geodesic} and incorporates an OT-based averaging step (c.f. \Cref{alg:average_point_clouds}). This adjustment preserves the algorithm's simplicity and scalability at a given refinement level, while effectively tackling the task of B-spline approximation. The defining characteristic of WLR is that the resulting curve is \emph{intrinsic} to the Wasserstein geometry since the geodesic averaging step stays within the Wasserstein space. This is similar to the computationally heavier algorithms of \cite{chen2018measure}, but different from e.g.\ \cite{chewi2021fast}, which is \textit{extrinsic} in the sense that it constructs splines in $\mathbb{R}^d$. An important consequence is that unlike prior work, WLR has the ability to automatically handle non-uniform mass, mass splitting and bifurcations.

\begin{algorithm}[t]
\caption{OT Averaging via Wasserstein Space Geodesic in \cref{eq:OT-geodesic}}
\label{alg:average_point_clouds}
\begin{algorithmic}[1]
\Procedure{OT-av}{$\mu_A, \mu_B, \alpha$} 
\State \textbf{input} Point cloud $\mu_A$: $(x_A, a_A)$
\State \textbf{input} Point cloud $\mu_B$: $(x_B, a_B)$
\State \textbf{input} Averaging parameter $\alpha \in [0, 1]$
\State $\gamma \gets \text{\Call{OT}{$\mu_A, \mu_B$}}$ 
\State \Comment{\textit{Solve \cref{eq:kantorovich} to obtain $\gamma \in \mathbb{R}^{n_A \times n_B}$}}
\State $x, a \gets $ null
\For{every $\gamma_{ij} > \epsilon$}
        \State Concatenate $(1 - \alpha) \cdot x_{A, i} + \alpha \cdot x_{B, j}$ to $x$
        \State Append $\gamma_{ij}$ to $a$

\EndFor
\State $a \gets a/$sum($a$) \Comment{\textit{Renormalization}}
\State \Return Point cloud $\mu_\alpha$ defined by $(x, a)$
\EndProcedure
\end{algorithmic}
\end{algorithm}

\begin{figure}[t]
    \centering
    \includegraphics[width=\linewidth]{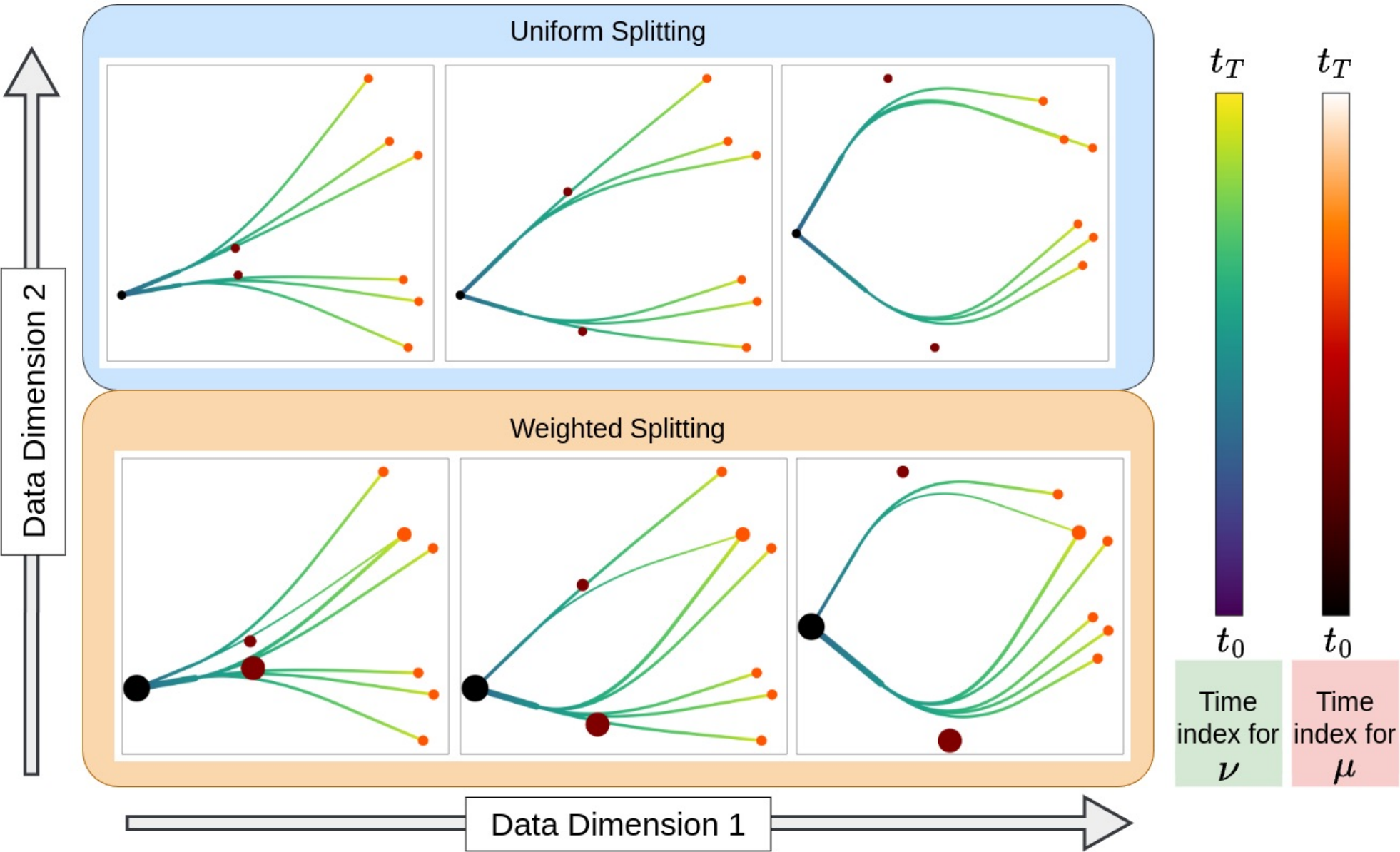}
    \caption{We provide an illustrative example in $\mathbb{R}^2$ to demonstrate how WLR infers trajectories and handles both uniform (Top) and non-uniform (Bottom) masses. The size of the point is proportional to $a_{{t_j}, i}$, the mass of point $i$ at time step $t_j$.  
    In all cases, the trajectories respect the geometry of the Wasserstein space.}
    \label{fig:splitting_fig}
\end{figure}

Before discussing numerical experiments, we illustrate our algorithm's behavior on two dimensional toy data.
\Cref{fig:splitting_fig} plots three point clouds ordered (in time) from left to right. In the top row, all points have uniform mass. In the bottom row, mass is non-uniformly distributed in the second and third time steps. This demonstrates the following capacities of WLR.

\paragraph{Trajectory inference} WLR produces natural trajectories between point clouds.
In our implementation of WLR, we repeat the end point clouds $\mu_{t_0}$ and $\mu_{t_T}$ before averaging (see (11) in \Cref{alg:WLRm}) so that they are interpolated. Therefore, when a sequence of point clouds $\nu$ is returned by WLR, we can set the first point cloud as $\nu_{t_0}= \mu_{t_0}$, the last point cloud as $\nu_{t_T}= \mu_{t_T}$, and evenly space out the remaining point clouds in between $t_0, t_1, \ldots, t_T$. In addition, if we compute OT between consecutive pairs of point clouds in $\nu$, we are able to trace how mass from a single point in $\mu_{t_0}$, i.e. mass of weight $a_{{t_0}, i}$ supported on $x_{{t_0}, i}$, travels from $t_0$ to $t_T$. In other words, we are able to trace trajectories of individual samples in $\mathbb{R}^d$ over time.

\paragraph{Approximation and bending behavior} B-splines in $\mathbb{R}^d$ do not interpolate given data. Rather, they find an approximating path between the knots.
\Cref{fig:splitting_fig} shows that this is still true for WLR and how the points in the second time step ``bend'' the trajectories like gravitational objects, and the bending is weighted by the mass located in each of the points.

\paragraph{Splitting trajectories} This phenomenon naturally occurs in point clouds with non-uniform weights. Our algorithm can accommodate this automatically through the geodesic averaging procedure.

Lastly, we note that OT-based geodesic averaging can be applied to subdivision schemes beyond B-splines. Following ideas from \cite{wallner2005subdivision}, we describe an interpolatory trajectory inference method based on the 4-point scheme by \cite{DYN1987pointscheme} in \Cref{sec:four-point}.

\subsection{Complexity analysis} \label{sec:complexity}
WLR's runtime depends on the B-spline degree $M$, number of initial point clouds $T+1$, refinement level $R$, and number of points per point cloud $n_t$. If the number of points stay the same, i.e. $n_t=n$, and there is a one-to-one mapping at each OT step, the overall complexity of WLR is $\gO \big( 2^R \cdot (T + M) \cdot n^3 \cdot \log n\big)$. Naturally, computation increases as $M$ (indicating smoothness) and $R$ (indicating approximation accuracy) increase. It is important to note that in the subdivision scheme literature, $R \le 8$ and $M=3$ suffice for most visual purposes. Still we provide a careful study of the interplay between these parameters for WLR runtime in \Cref{sec:runtime}. 

 The most time consuming procedure is computing discrete OT in (5) of \Cref{alg:average_point_clouds}, for which we use linear programming implemented in Python OT (\cite{PythonOT}).  In practice, we also vectorize lines (7-14) in \Cref{alg:average_point_clouds}. Since the averaging operations only happen between consecutive point clouds, it is theoretically possible to parallelize WLR, but we leave that for future work.

\subsection{Convergence analysis}\label{sec:convergence}

For the purpose of the convergence proof, we consider the data $\mu_{t_j},j=0,\ldots, T$ in the $p$-Wasserstein space over $\mathbb{R}^d$, $\mathcal{W}_p(\mathbb{R}^d)$, which is the space of all probability measure with finite $p$-th moment. This space is a metric space with distance given by the $p$-Wasserstein distance $W_p$ in \Cref{sec:OT}.

The $p$-Wasserstein distance satisfies the geodesic property: for any two measures, $\mu_i$ and $\mu_j$ we have for $\alpha \in [0,1]$,
\begin{equation} \label{eqn:geodesic_property}
	W_p\left(\operatorname{OT-av}_{\alpha}(\mu_i,\mu_j),\mu_j\right) = (1-\alpha) \, W_p(\mu_i,\mu_j).
\end{equation}
In addition, $\mathcal{W}_p(\mathbb{R}^d)$ is a separable complete metric space since $\mathbb{R}^d$ has these properties (\cite{villani2008optimal}). Therefore, any Cauchy sequence in $\mathcal{W}_p(\mathbb{R}^d)$ converges, which is the main property we need for the following convergence result.

Our subdivision scheme WLR generates sequences of point clouds, starting from the initial data $\nu^{(0)} = [\nu^{(0)}_{t_j}]_{j=0}^T$ with $\nu^{(0)}_{t_j} = \mu_{t_j}$, and generating, by repeated refinements, the sequences $\nu^{(R)} = [\nu^{(R)}_j]_{j=0}^{2^RT+1}$, $R \in \mathbb{N}$. Therefore, informally, the limit $\nu = \lim_{R\to \infty}\nu^{(R)}$ is the object we seek in \Cref{sec:formulation}. Technically, we associate each refinement level's data $[ (j2^{-R},\nu^{(R)}_j)]_{j=0}^{2^RT+1}$ with the piecewise geodesic interpolant in $\mathcal{W}_p(\mathbb{R}^d)$, which is defined using the Wasserstein geodesic of \cref{eq:OT-geodesic} and is calculated by 
\Cref{alg:average_point_clouds}. The piecewise geodesic interpolant is 
\[ N^{(R)}(t) = \operatorname{OT-av}_{t}(\nu^{(R)}_j, \nu^{(R)}_{j+1}), \quad t \in [j2^{-R},(j+1)2^{-R}). \] Then, the refinement scheme (e.g.\ WLR) is called convergent if the sequence of piecewise geodesic interpolants $\{N^{(R)}(t)\}_{R\in\mathbb{Z}_+}$ converges uniformly (in $t$) to a limit curve, where the curve takes values in the Wasserstein space $\mathcal{W}_p(\mathbb{R}^d)$.  This is called the \emph{limit of the subdivision scheme}. 

The result for our WLR schemes reads:
\begin{thm}\label{thm:convergence-subdiv}
Consider a sequence of measures $\mu_{t_0},\ldots,\mu_{t_T}$ with $t_0<\ldots < t_T$. Define the initial data by $\nu^{(0)} = [\nu^{(0)}_{t_j}]_{j=0}^T$ with $\nu^{(0)}_{t_j} = \mu_{t_j}$, and choose a smoothness degree $M$. Apply WLR (i.e.\ call \Cref{alg:WLRm} for the initial data, smoothness $M$, and refinement level $R$) to obtain a $R$-times refined sequence of measures $\nu^{(R)}$. As $R\to \infty$, WLR converges to a continuous limit curve $\nu$ in $\mathcal{W}_p(\mathbb{R}^d)$, where convergence is in the sense of subdivision schemes defined above. 

For the obtained piecewise geodesic connecting the sequence of measures in the $R$-th iteration step $(N^{(R)}(t))$, we obtain a linear convergence rate:
\begin{equation} \label{eqn:thm1}
    W_p(N^{(\infty)}(t), N^{(R)}(t)) \leq C \frac{1}{2^{R-1}} = \mathcal{O}(2^{-R}),
\end{equation}
where constant $C$ is a constant independent of $R$.

\end{thm}

\begin{proof}
The proof follows from adapting the results of \cite[Corollary 3.3]{sharon2017global} for finite-dimensional Riemannian manifolds to the Wasserstein space. 


First, we recall the notion 
\begin{equation} \label{eqn:deltaP}
	\Delta \nu^{(R)} := \sup_{j\in\mathbb{Z}} W_p(\nu^{(R)}_j,\nu^{(R)}_{j+1}) ,
\end{equation} 
and assume that the initial data is bounded, that is, $ \Delta \nu^{(0)} < \infty $. Following the arguments in the proof of \cite{sharon2017global}, we get that there exist $\eta \in (0,1)$ such that
\[ \Delta \nu^{(R)} \le \eta \Delta \nu^{(R-1)} , \quad R \in \mathbb{N}.  \]
We term the factor $\eta$ \emph{contractivity factor}. Using the geodesic property of \cref{eqn:geodesic_property}, one can actually show that WLR has a contractivity factor of $\eta=1/2$. In fact, this value equals the contractivity factor of the original linear Lane-Riesenfeld scheme, which is the optimal contractivity for linear subdivision schemes, see \cite {Dyn2024convergence}.

To follow the proof of \cite{sharon2017global}, we also need an additional mild condition, the \emph{displacement-safe property}, which is immediately satisfied for WLR, because it is satisfied for the standard linear Lane-Riesenfeld algorithm.
This, together with the contractivity property, implies that there exists a constant $C$, independent of $R$, such that $W_p(N^{(R+1)}(t), N^{R}(t))\le C \eta^R$. And so the convergence follows, since for any $\ell \in \mathbb{N}$, we have 

\begin{align*}
W_p(N^{(R+\ell)}(t), N^{R}(t)) &\le C \eta^R (1+\eta+\ldots + \eta^\ell) \\
&\le \frac{C}{1-\eta} \eta^R.
\end{align*}
Namely, $\{N^{(R)}(t)\}_{R\in\mathbb{Z}_+}$ is a Cauchy sequence in the Wasserstein space, and since this space is complete, the sequence converges to a limit.
By letting $\ell \to \infty$, we obtain a bound on the difference between the limit curve and the $R$-th iteration:
\begin{equation*}
    W_p(N^{(\infty)}(t), N^{(R)}(t)) \leq \frac{C}{2^{R-1}} = \mathcal{O}(2^{-R}). \qedhere
\end{equation*}

\end{proof}

\section{EXTENSION OF WLR AND THE FOUR-POINT SCHEME}\label{sec:four-point}

Although the Wasserstein Lane–Riesenfeld algorithm (\Cref{alg:WLRm}, an extension of \Cref{alg:LRm}) is our primary focus, the approach introduced in \Cref{sec:method} naturally applies to any subdivision scheme. In particular, replacing each linear average with an \(\operatorname{OT-av}\) results in a broad family of both interpolatory and approximating schemes in Wasserstein space. We choose WLR to illustrate this idea because it can be tuned to different smoothness levels through the parameter \(M\), and it exhibits strong convergence properties \cite{sharon2017geodesic,sharon2017global}, which we extend in \Cref{sec:convergence}.

\begin{figure}[H]
    \centering
    \includegraphics[width=\linewidth]{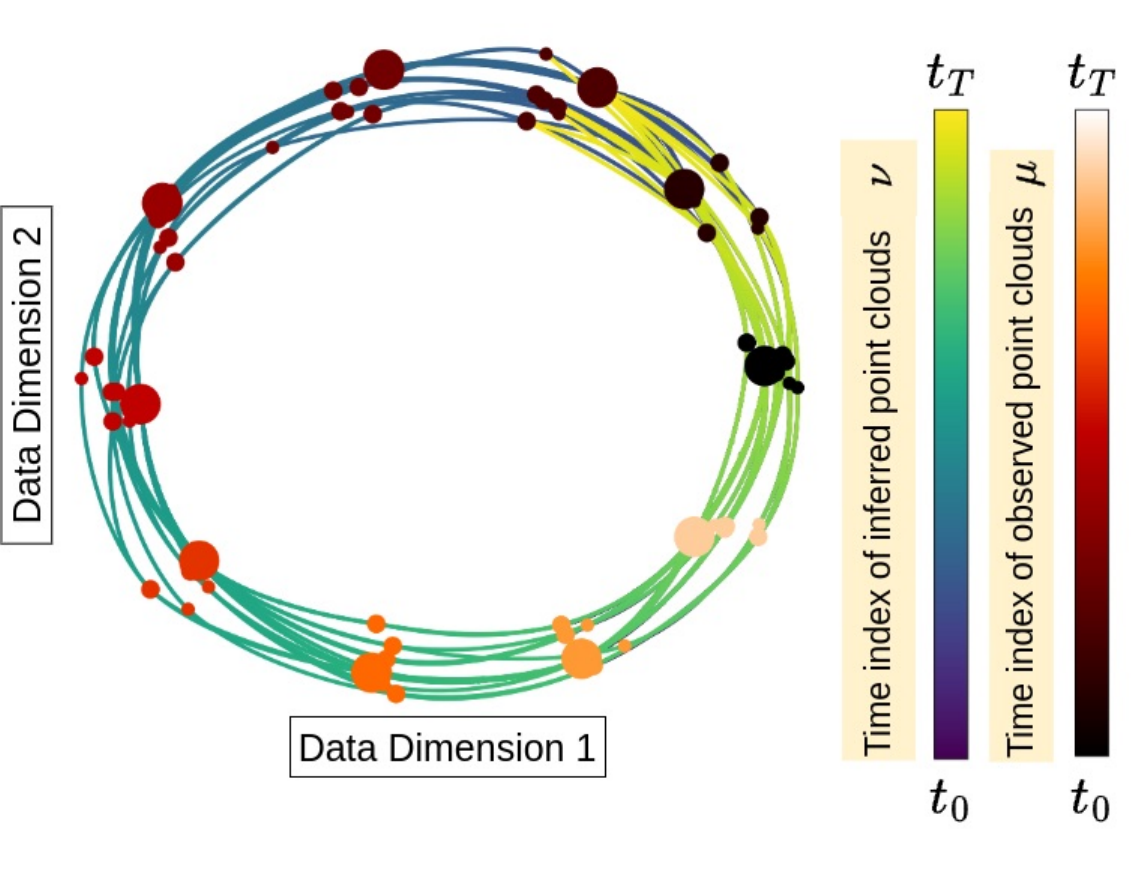}
    \caption{Illustrative example in $\mathbb{R}^2$ demonstrating exact interpolation via the Wasserstein four-point scheme on weighted circular Gaussians.}
    \label{fig:weighted_circle_four_point}
\end{figure}

As an interpolatory alternative, we also adapt the classical four-point scheme of \cite{DYN1987pointscheme} by substituting all linear averages with \(\operatorname{OT-av}\). The resulting Wasserstein four-point scheme is described in \Cref{alg:PC4PointScheme} and demonstrated in \Cref{fig:weighted_circle_four_point} on two-dimensional circular, non-uniform Gaussian data. Notably, its contractivity factor is \(\eta = 4\omega + \tfrac12\) \cite{dyn2017manifold}. For the parameter \(\omega = \tfrac{1}{16}\), we have \(\eta = \tfrac{3}{4}\). By an argument analogous to the convergence proof of WLR in \Cref{sec:convergence}, this contractivity guarantees convergence of the four-point scheme in Wasserstein space.

\begin{algorithm}[htp]
    \caption{4-Point Scheme in Wasserstein Space}
    \label{alg:PC4PointScheme}
    \begin{algorithmic}[1]
    \Procedure{FourPointScheme}{$[\mu_{t_j}]_{j=0}^T, R$}
        \State \textbf{input} Point clouds to be refined $ [\mu_{t_j}]_{j=0}^T$.
        \State \textbf{input} Refinement level $R \in \mathbb{Z}_{+}$.
        \State $\nu^{(0)} \gets [\mu_{t_j}]_{j=0}^T$
        \State $w \gets \frac{1}{16}$ \Comment{\textit{$w$ is the weighting parameter}}
        \For{$r = 1$ \textbf{to} $R$}
            \For{$j = 2$ \textbf{to} $(|\nu^{(r-1)}| - 1)$}
                \State $\nu^{(r-1)}_{2j} \gets \nu^{(r-1)}_j$
            \EndFor
            \For{$j = 2$ \textbf{to} $(|\nu^{(r-1)}| - 2)$}
                \State $X_A \gets \Call{OT-av}{\nu_j^{(r-1)}, \nu_{j-1}^{(r-1)}, -2w}$
                \State $X_B \gets \Call{OT-av}{\nu_{j+1}^{(r-1)}, \nu_{j+2}^{(r-1)}, -2w}$
                \State $\nu_{2j+1}^{(r)} \gets \Call{OT-av}{X_A, X_B, \frac{1}{2}}$
            \EndFor
            \State $\nu^{(r)} \gets [\nu_{4}^{(r)}, \ldots, \nu_{2(|\nu^{(r-1)}|-1)}^{(r)}]$
        \EndFor
        \State \Return Refined point clouds $\nu^{(R)}$.
    \EndProcedure
    \end{algorithmic}
\end{algorithm}

\section{EXPERIMENTS}\label{sec:experiments}

\Cref{sec:supercells}  highlights settings where mass splitting occurs or where point clouds across time steps have different number of points, which WLR is uniquely suited for. We then evaluate the performance of WLR against other trajectory inference methods on uniformly distributed mass in \Cref{sec:sota-comparison}.
In both settings, performance is evaluated with the leave-one-out experiment as suggested in \cite{huguet2022manifold}. For each algorithm, we withhold data at a common intermediate time step during the `training' phase. Then, in the `inference' phase, we predict the outcomes for the omitted time step and document the 1-Wasserstein ($W_1$) distance between the predicted and the omitted point clouds. We compared WLR against six state-of-the-art algorithms: MIOFlow (\cite{huguet2022manifold}), TrajectoryNet (\cite{tong2020trajectorynet}), Conditional Flow Matching (\cite{CFM_tong2023improving}), F \& S algorithm (\cite{chewi2021fast}), Wasserstein-Fisher-Rao (\cite{clancy2022wassersteinfisherrao}), and Wasserstein Lagrangian Flow (\cite{wlflow_neklyudov_computational_2023}). We repeated each experiment by subsampling within each dataset using five different random seeds. 
When possible, we optimized code and tuned parameters of other methods for fair comparison.  
See \Cref{app:experiment} for all experimental setup and dataset details. Our Python code is available at \url{https://github.com/amartya21/Wasserstein-Trajectory-Inference}.

\subsection{Splitting trajectories and non-uniform weights in supercells} \label{sec:supercells}

In this section, we designed two experiments to demonstrate the effectiveness of WLR in handling these mass splitting scenarios. The first uses the Converging Gaussian data by \cite{clancy2022wassersteinfisherrao}, which has different number of points with uniformly distributed mass per time step.
The inferred trajectories from the use of all time steps in the data sets are visualized in \Cref{fig:failure_cases}, and the leave-one-out results are reported in \Cref{tab:conv-gauss}, in which WLR performs best or second best numerically.

\begin{table}[htp]
\centering
\scriptsize
\setlength{\tabcolsep}{3pt}
\begin{tabular}{lccc}
\toprule
\multirow{2}{*}{\textbf{Method}} 
 & \textbf{Runtime} & \textbf{Leave-one-out} & \textbf{Mean} \\
 & \textbf{(sec) $\downarrow$} & $\mathbf{W_1}\downarrow$ & $\mathbf{W_1}\downarrow$ \\
\midrule
WLR (Ours)
  & $63.50 \pm 118.3$
  & $\mathbf{0.70 \pm 0.0}$
  & $\underline{0.29 \pm 0.3}$ \\
WFR
  & $\mathbf{2.72 \pm 0.85}$
  & $0.92 \pm 0.24$
  & $0.44 \pm 0.13$ \\
CFM
  & $167.84 \pm 9.37$
  & $\underline{0.72 \pm 0.01}$
  & $\mathbf{0.27 \pm 0.29}$ \\
MIOFlow
  & $\underline{32.25 \pm 0.37}$
  & $2.78 \pm 2.64$
  & $2.20 \pm 1.89$ \\
TrajectoryNet
  & $1310.13 \pm 55.54$
  & $1.94 \pm 0.03$
  & $2.22 \pm 0.51$ \\
WLF-SB
  & $92.87 \pm 10.10$
  & $25.64 \pm 19.04$
  & $25.61 \pm 18.89$ \\
WLF-UBOT+
  & $78.94 \pm 9.00$
  & $0.95 \pm 0.07$
  & $1.04 \pm 0.05$ \\
\bottomrule
\end{tabular}
\caption{WLR performs the \textbf{best} or \underline{second best} on Converging Gaussian, which has a different number of points per time step with uniformly distributed mass. See Figure \ref{fig:failure_cases} for corresponding visualizations.}
\label{tab:conv-gauss}
\end{table}

The second experiement is on CITE-Seq supercells, where the mass distribution is not uniform. In single-cell genomics, the concept of \textit{supercells} or \textit{metacells} has gained popularity in recent years as a way to simplify scRNA-seq data (\cite{supercell_baran2019metacell, supercell_chen2021single, supercell_bilous2022metacells}). Supercells are formed by merging transcriptionally similar cells by creating a $k$-nearest neighbor graph based on gene expression. By aggregating similar cells, supercells can mitigate the effects of technical noise and dropout events, provide a more robust representation of the underlying biological state, and capture rare cell types or transient states that might be overlooked in individual cell analysis. This simplification aids in downstream analyses, such as visualization and differential expression analysis and allows for a more comprehensive understanding of the cellular heterogeneity within a sample.

Creation of supercells naturally gives rise to data with different numbers of points per time step, i.e. $n_t$ that changes over $t$, which then leads to splitting trajectories. 
Existing trajectory inference methods struggle to accurately model such scenarios, as they typically rely on unbalanced transport and assume uniformly distributed mass at each time step. These assumptions do not adequately capture the complexity of real-world systems where masses can split unevenly or follow diverse paths. For example, it is intuitive to assign weights to supercells such that the weights are proportional to the number of cells that have been merged to create each supercell.

\begin{figure}[htp]
    \centering
    \includegraphics[width=1\linewidth]{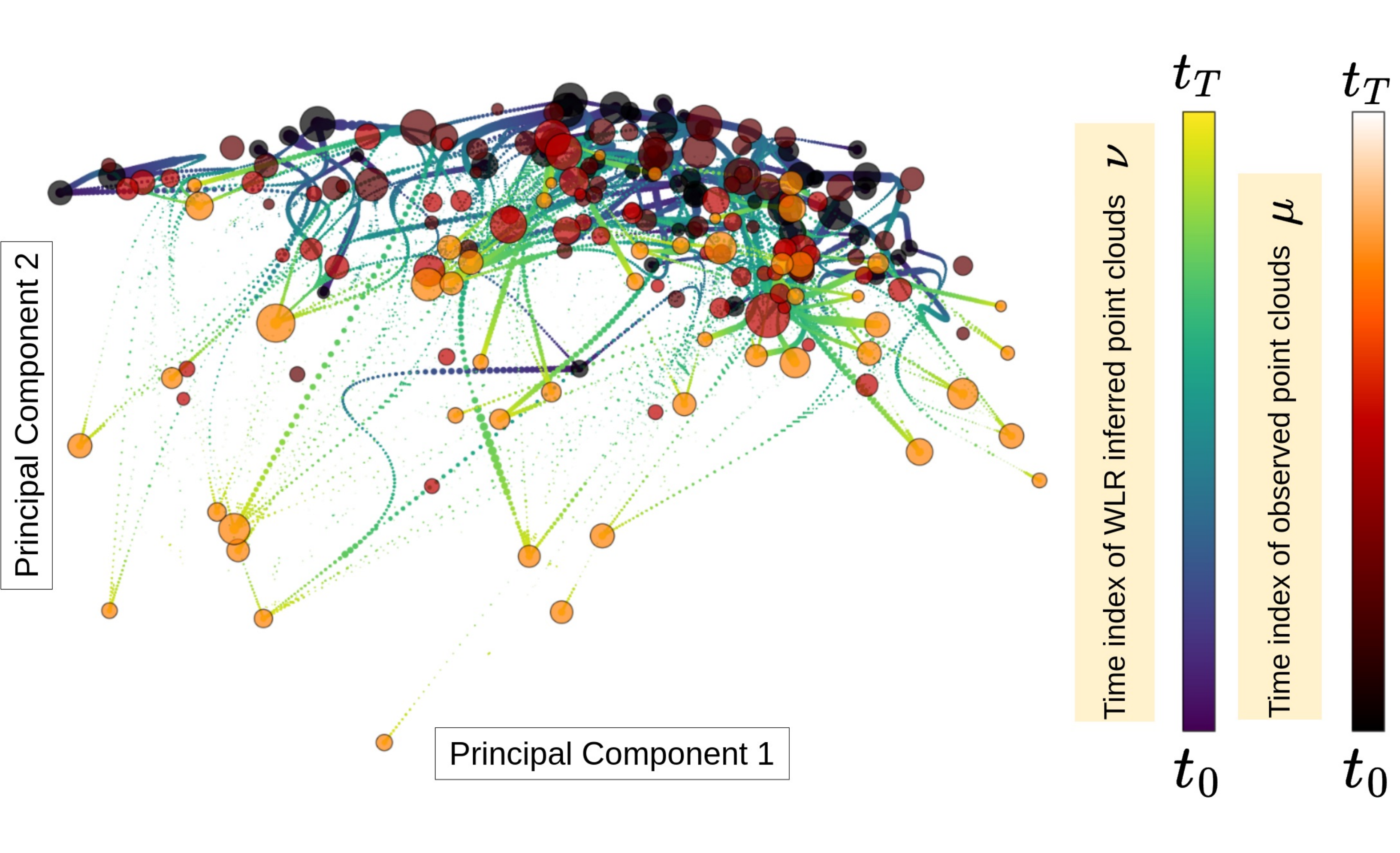}
    \caption{WLR produces smooth trajectories on CITE-seq supercells with non-uniformly distributed mass by automatically splitting trajectories. The size of the points is proportional to $a_{{t_j}, i}$, the mass of point $i$ at time step $t_j$. Only a subset $(n_t = 50)$ is visualized for clarity. }
    \label{fig:citeseq-supercells}
\end{figure}

Existing trajectory inference methods typically struggle with such scenarios as they rely on unbalanced transport and assume uniform mass distributions. To the best of our knowledge, there are currently no neural-network based or spline-based methods in the literature that address trajectory inference with weighted mass distributions by accounting for splitting. As a result, \Cref{fig:citeseq-supercells} and \Cref{tab:loo_supercell_wasserstein} report standalone values.

\begin{table}[htp]
    \centering
    \scriptsize
      \footnotesize
    \begin{tabular}{c c c}
    \toprule
    \textbf{$R$} & \textbf{Time} & \textbf{Leave-one-out $W_1 \downarrow$} \\
    \midrule
    4 & 25.17 mins & 23.90701 \\
    5 & 5.77 hours & 23.56288 \\
    6 & 1.19 days & 21.49304 \\
    \bottomrule
    \end{tabular}
    \caption{WLR is capable of trajectory inference on CITE-seq supercells at varying refinement levels.}
    \label{tab:loo_supercell_wasserstein}
\end{table}

\subsection{Constant number of points with uniform mass distribution}\label{sec:sota-comparison}

This section applies WLR to the setting where the number of points $n_t$ stays constant with uniform mass distribution. 
Our primary objective in these experiments is not strictly to surpass the existing state-of-the-art methods but rather to demonstrate that our approach is competitive and provides distinct advantages when dealing with more complex scenarios such as in the case of supercells.

 We use a wide range of simulated and real datasets for this task: Diverging Gaussian, Petal (\cite{huguet2022manifold}), Dyngen Tree and Cycle (\cite{dyngen_cannoodt2021}), and CITE-seq (\cite{citeseq_stoeckius2017large}). These datasets capture the natural dynamics observed in cellular differentiation such as bifurcations and merges.

\begin{figure}[htp]
    \centering
    \includegraphics[width=\linewidth]{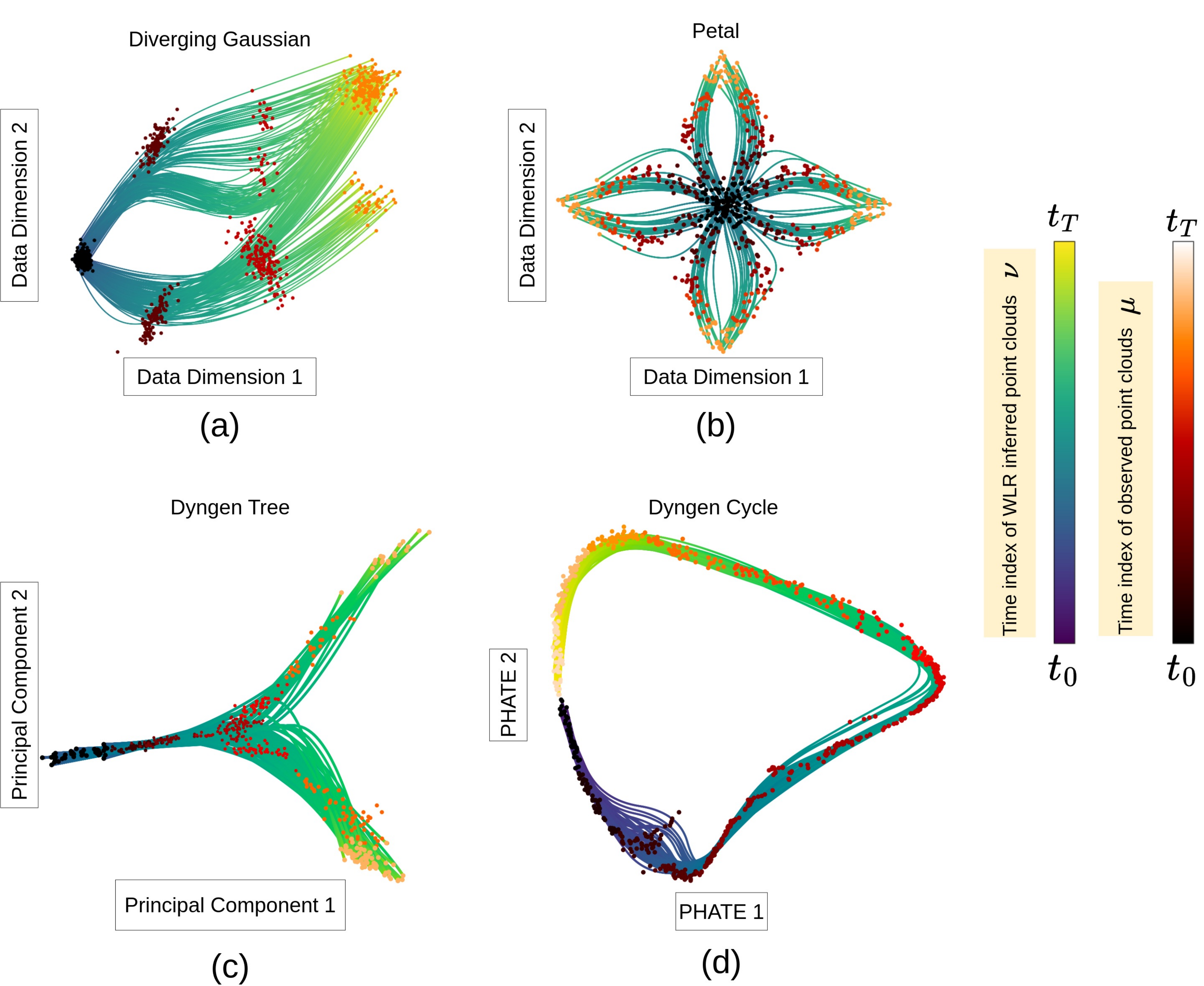}
    \caption{WLR produces smooth and visually convincing trajectories on four synthetic datasets that mimic cell differentiation.}
    \label{fig:WLR_interp}
\end{figure}

 These experiments are split up into two categories. In the first set of results in \Cref{tab:experiments}, we compare B-spline based approximation methods and neural network approaches, both of which provide smooth approximations rather than exact interpolations of the input point clouds. We also perform trajectory inference using all time steps in the datasets and visualize WLR trajectories in \Cref{fig:WLR_interp}; visualization of other methods are deferred to \Cref{fig:full_scale_experiemnts} in \Cref{app:additional_results}. 
 In the second set of results in \Cref{tab:interpolation_experiments}, we compare the interpolatory methods for trajectory inference, namely the original F \& S method proposed by \cite{chewi2021fast} and our four point scheme described in \Cref{alg:PC4PointScheme}.

\subsection{Discussion}
Our experiments demonstrate that WLR performs consistently well across diverse datasets in terms of runtime, leave-one-out $W_1$ distance, leave-one-out MSE, and qualitative trajectory smoothness. WLR can dependably model complex dynamics such as bifurcations, merges, and trajectory splitting with both uniform and non-uniform weights.
In comparison, deep learning methods tend to require long training times and exhibit optimization instability. 
Deep learning-based methods also have inherent limitations when it comes to handling mass splitting phenomena. These approaches maintain a fixed number of trajectories from initialization to termination, determined by the initial configuration, making it challenging to naturally model scenarios where particle masses split or merge. The only way these methods can approximate mass splitting behavior is by working in the unbalanced optimal transport setting. However, this is an indirect workaround that does not fully capture the underlying splitting dynamics. In contrast, our method can directly handle mass splitting phenomena without requiring such approximations, making it more suitable for applications where trajectory branching is a natural part of the system dynamics.
Note that quantitative performance does not always align with qualitative plausibility of the trajectories; see \Cref{fig:full_scale_experiemnts} and \Cref{fig:failure_cases}.
F\&S produces results similar to---often identical to---WLR and offers faster runtimes. However, F\&S is not designed to handle trajectory splitting or varying number of points across time steps.
The Wasserstein-Fisher-Rao splines improve upon F\&S via unbalanced transport, but optimizing for a different geometry can lead to crossing trajectories or omitted time steps.

\section{CONCLUSIONS}
We present a simple framework for trajectory inference on point clouds that respects the inherent geometry in the data. Our method builds on subdivision scheme methods with optimal transport-based geodesic to efficiently and provably approximate (or interpolate) curves in the Wasserstein space. Unlike prior work based on neural ODEs, which may fail under stiff dynamics or rely heavily on hyperparameters, our method consistently provides smooth and accurate trajectories in a user-controlled way. While our current implementation of WLR uses a straightforward linear programming approach to solve OT problems sequentially, we believe that parallelization could substantially reduce runtime. Since WLR requires solving OT only between consecutive time steps, these subproblems could, in principle, be run in parallel, although special care would be needed at the boundaries where sub-trajectories meet. In future work, improving scalability and efficiency could greatly extend WLR’s applicability to large-scale datasets. We demonstrate WLR's capability to handle high-dimensional complex biological datasets like CITE-seq supercells and show the ability to carry out trajectory inference in uniform and weighted mass splitting scenarios.
Moreover, based on our preliminary results in \Cref{sec:NN_initialization}, we posit that employing our framework as an initialization step could enhance the convergence rates of various neural network-based trajectory inference methods.

\subsection{Acknowledgements}
We thank the authors of \cite{chewi2021fast} and \cite{justiniano2023approximation} for sharing their code with us. We also thank Natalie Stanley for helpful discussions and introducing us to supercells. NS is partially supported by the NSF-BSF award 2019752 and the DFG award 514588180.
CM is partially supported by NSF awards DMS-2306064 and DMS-2410140, and by a seed grant from the School of Data Science and Society at UNC. There are no competing interests to declare.

\bibliography{main_aistats}

\begin{thebibliography}{}

\bibitem[Agueh and Carlier, 2011]{agueh2011barycenters}
Agueh, M. and Carlier, G. (2011).
\newblock Barycenters in the wasserstein space.
\newblock {\em SIAM Journal on Mathematical Analysis}, 43(2):904--924.

\bibitem[Ambrosio et~al., 2013]{ambrosio2013user}
Ambrosio, L., Bressan, A., Helbing, D., Klar, A., Zuazua, E., Ambrosio, L., and
  Gigli, N. (2013).
\newblock A user’s guide to optimal transport.
\newblock {\em Modelling and Optimisation of Flows on Networks: Cetraro, Italy
  2009, Editors: Benedetto Piccoli, Michel Rascle}, pages 1--155.

\bibitem[Baccou and Liandrat, 2024]{baccou2024subdivision}
Baccou, J. and Liandrat, J. (2024).
\newblock Subdivision scheme for discrete probability measure-valued data.
\newblock {\em Applied Mathematics Letters}, 158:109233.

\bibitem[Baran et~al., 2019]{supercell_baran2019metacell}
Baran, Y., Bercovich, A., Sebe-Pedros, A., Lubling, Y., Giladi, A., Chomsky,
  E., Meir, Z., Hoichman, M., Lifshitz, A., and Tanay, A. (2019).
\newblock Metacell: analysis of single-cell rna-seq data using k-nn graph
  partitions.
\newblock {\em Genome biology}, 20:1--19.

\bibitem[Benamou et~al., 2019]{benamou2019cubicsplines}
Benamou, J.-D., Gallou\"et, T.~O., and Vialard, F.-X. (2019).
\newblock Second-order models for optimal transport and cubic splines on the
  {W}asserstein space.
\newblock {\em Found Comp Math}, 19:113--1143.

\bibitem[Bilous et~al., 2022]{supercell_bilous2022metacells}
Bilous, M., Tran, L., Cianciaruso, C., Gabriel, A., Michel, H., Carmona, S.~J.,
  Pittet, M.~J., and Gfeller, D. (2022).
\newblock Metacells untangle large and complex single-cell transcriptome
  networks.
\newblock {\em BMC bioinformatics}, 23(1):336.

\bibitem[Cannoodt et~al., 2021]{dyngen_cannoodt2021}
Cannoodt, R., Saelens, W., Deconinck, L., and Saeys, Y. (2021).
\newblock Spearheading future omics analyses using dyngen, a multi-modal
  simulator of single cells.
\newblock {\em Nature Communications}, 12(1):3942.

\bibitem[Cavaretta et~al., 1989]{cavaretta1989stationary}
Cavaretta, A., Dahmen, W., and Micchelli, C. (1989).
\newblock {\em Stationary Subdivision}.
\newblock Preprint / A: Fachbereich Mathematik. Freie Universit{\"a}t Berlin.
  Fachbereich Mathematik.

\bibitem[Chen et~al., 2021]{supercell_chen2021single}
Chen, L., He, Q., Zhai, Y., and Deng, M. (2021).
\newblock Single-cell rna-seq data semi-supervised clustering and annotation
  via structural regularized domain adaptation.
\newblock {\em Bioinformatics}, 37(6):775--784.

\bibitem[Chen et~al., 2018]{chen2018measure}
Chen, Y., Conforti, G., and Georgiou, T.~T. (2018).
\newblock Measure-valued spline curves: An optimal transport viewpoint.
\newblock {\em SIAM Journal on Mathematical Analysis}, 50(6):5947--5968.

\bibitem[Chewi et~al., 2021]{chewi2021fast}
Chewi, S., Clancy, J., Le~Gouic, T., Rigollet, P., Stepaniants, G., and
  Stromme, A. (2021).
\newblock Fast and smooth interpolation on {W}asserstein space.
\newblock In {\em International Conference on Artificial Intelligence and
  Statistics}, pages 3061--3069. PMLR.

\bibitem[Clancy and Suarez, 2022]{clancy2022wassersteinfisherrao}
Clancy, J. and Suarez, F. (2022).
\newblock Wasserstein-{F}isher-{R}ao splines.
\newblock arXiv:2203.15728.

\bibitem[Courty et~al., 2017]{courty2017joint}
Courty, N., Flamary, R., Habrard, A., and Rakotomamonjy, A. (2017).
\newblock Joint distribution optimal transportation for domain adaptation.
\newblock {\em Advances in neural information processing systems}, 30.

\bibitem[Craig et~al., 2024]{craig2024blob}
Craig, K., Elamvazhuthi, K., and Lee, H. (2024).
\newblock A blob method for mean field control with terminal constraints.
\newblock {\em arXiv preprint arXiv:2402.10124}.

\bibitem[Dyn et~al., 1987]{DYN1987pointscheme}
Dyn, N., Levin, D., and Gregory, J.~A. (1987).
\newblock A 4-point interpolatory subdivision scheme for curve design.
\newblock {\em Computer Aided Geometric Design}, 4(4):257--268.

\bibitem[Dyn and Sharon, 2017a]{sharon2017global}
Dyn, N. and Sharon, N. (2017a).
\newblock A global approach to the refinement of manifold data.
\newblock {\em Mathematics of Computation}, 86(303):pp. 375--395.

\bibitem[Dyn and Sharon, 2017b]{sharon2017geodesic}
Dyn, N. and Sharon, N. (2017b).
\newblock Manifold-valued subdivision schemes based on geodesic inductive
  averaging.
\newblock {\em Journal of Computational and Applied Mathematics}, 311:54--67.

\bibitem[Dyn and Sharon, 2017c]{dyn2017manifold}
Dyn, N. and Sharon, N. (2017c).
\newblock Manifold-valued subdivision schemes based on geodesic inductive
  averaging.
\newblock {\em Journal of Computational and Applied Mathematics}, 311:54--67.

\bibitem[Dyn and Sharon, 2024]{Dyn2024convergence}
Dyn, N. and Sharon, N. (2024).
\newblock Improving the convergence analysis of linear subdivision schemes.
\newblock {\em arXiv preprint arXiv:2405.09414}.

\bibitem[Flamary et~al., 2021]{PythonOT}
Flamary, R., Courty, N., Gramfort, A., Alaya, M.~Z., Boisbunon, A., Chambon,
  S., Chapel, L., Corenflos, A., Fatras, K., Fournier, N., et~al. (2021).
\newblock Pot: Python optimal transport.
\newblock {\em Journal of Machine Learning Research}, 22(78):1--8.

\bibitem[Howe et~al., 2022]{howe2022myriad}
Howe, N., Dufort-Labb{\'e}, S., Rajkumar, N., and Bacon, P.-L. (2022).
\newblock Myriad: a real-world testbed to bridge trajectory optimization and
  deep learning.
\newblock {\em Advances in Neural Information Processing Systems},
  35:29801--29815.

\bibitem[Huang et~al., 2022]{huang2022representation}
Huang, X., Wang, Y., Guizilini, V., Ambrus, R., Gaidon, A., and Solomon, J.
  (2022).
\newblock Representation learning for object detection from unlabeled point
  cloud sequences.
\newblock In {\em Conference on Robot Learning}.

\bibitem[Huguet et~al., 2022]{huguet2022manifold}
Huguet, G., Magruder, D.~S., Tong, A., Fasina, O., Kuchroo, M., Wolf, G., and
  Krishnaswamy, S. (2022).
\newblock Manifold interpolating optimal-transport flows for trajectory
  inference.
\newblock {\em Advances in Neural Information Processing Systems},
  35:29705--29718.

\bibitem[Justiniano et~al., 2024]{justiniano2023approximation}
Justiniano, J., Rumpf, M., and Erbar, M. (2024).
\newblock Approximation of splines in wasserstein spaces.
\newblock {\em ESAIM: Control, Optimisation and Calculus of Variations}, 30:64.

\bibitem[Kantorovich, 2006]{Kantorovich2006_OptimalTransport}
Kantorovich, L.~V. (2006).
\newblock On the translocation of masses.
\newblock {\em Journal of Mathematical Sciences}, 133(4):1381--1382.

\bibitem[Lane and Riesenfeld, 1980]{Lane1980LRalgorithm}
Lane, J.~M. and Riesenfeld, R.~F. (1980).
\newblock A theoretical development for the computer generation and display of
  piecewise polynomial surfaces.
\newblock {\em IEEE Transactions on Pattern Analysis and Machine Intelligence},
  PAMI-2(1):35--46.

\bibitem[Lott, 2008]{LottWassersteinRiemannian2008}
Lott, J. (2008).
\newblock Some geometric calculations on {W}asserstein space.
\newblock {\em Comm. Math. Phys.}, 277:423--437.

\bibitem[Moon et~al., 2019]{phate_moon2019visualizing}
Moon, K.~R., Van~Dijk, D., Wang, Z., Gigante, S., Burkhardt, D.~B., Chen,
  W.~S., Yim, K., Elzen, A. v.~d., Hirn, M.~J., Coifman, R.~R., et~al. (2019).
\newblock Visualizing structure and transitions in high-dimensional biological
  data.
\newblock {\em Nature biotechnology}, 37(12):1482--1492.

\bibitem[Neklyudov et~al., 2023]{wlflow_neklyudov_computational_2023}
Neklyudov, K., Brekelmans, R., Tong, A., Atanackovic, L., Liu, Q., and
  Makhzani, A. (2023).
\newblock A computational framework for solving wasserstein lagrangian flows.

\bibitem[Otto, 2001]{OttoRiemannianOT2001}
Otto, F. (2001).
\newblock The geometry of dissipative evolution equations: the porous medium
  equation.
\newblock {\em Comm. Partial Differential Equations}, 26(1-2):101--174.

\bibitem[Peyr{\'e} et~al., 2019]{peyre2019computational}
Peyr{\'e}, G., Cuturi, M., et~al. (2019).
\newblock Computational optimal transport: With applications to data science.
\newblock {\em Foundations and Trends{\textregistered} in Machine Learning},
  11(5-6):355--607.

\bibitem[Saelens et~al., 2019]{saelens2019comparison}
Saelens, W., Cannoodt, R., Todorov, H., and Saeys, Y. (2019).
\newblock A comparison of single-cell trajectory inference methods.
\newblock {\em Nature biotechnology}, 37(5):547--554.

\bibitem[Santambrogio, 2015]{santambrogio2015optimal}
Santambrogio, F. (2015).
\newblock Optimal transport for applied mathematicians.
\newblock {\em Birk{\"a}user, NY}, 55(58-63):94.

\bibitem[Schiebinger et~al., 2019]{schiebinger2019optimal}
Schiebinger, G., Shu, J., Tabaka, M., Cleary, B., Subramanian, V., Solomon, A.,
  Gould, J., Liu, S., Lin, S., Berube, P., et~al. (2019).
\newblock Optimal-transport analysis of single-cell gene expression identifies
  developmental trajectories in reprogramming.
\newblock {\em Cell}, 176(4):928--943.

\bibitem[Sha et~al., 2023]{sha2023reconstructing}
Sha, Y., Qiu, Y., Zhou, P., and Nie, Q. (2023).
\newblock Reconstructing growth and dynamic trajectories from single-cell
  transcriptomics data.
\newblock {\em Nature Machine Intelligence}, pages 1--15.

\bibitem[Stoeckius et~al., 2017]{citeseq_stoeckius2017large}
Stoeckius, M., Hafemeister, C., Stephenson, W., Houck-Loomis, B.,
  Chattopadhyay, P.~K., Swerdlow, H., Satija, R., and Smibert, P. (2017).
\newblock Large-scale simultaneous measurement of epitopes and transcriptomes
  in single cells.
\newblock {\em Nature methods}, 14(9):865.

\bibitem[Tong et~al., 2024]{CFM_tong2023improving}
Tong, A., FATRAS, K., Malkin, N., Huguet, G., Zhang, Y., Rector-Brooks, J.,
  Wolf, G., and Bengio, Y. (2024).
\newblock Improving and generalizing flow-based generative models with
  minibatch optimal transport.
\newblock {\em Transactions on Machine Learning Research}.
\newblock Expert Certification.

\bibitem[Tong et~al., 2020]{tong2020trajectorynet}
Tong, A., Huang, J., Wolf, G., Van~Dijk, D., and Krishnaswamy, S. (2020).
\newblock Trajectory{N}et: A dynamic optimal transport network for modeling
  cellular dynamics.
\newblock In {\em International conference on machine learning}, pages
  9526--9536. PMLR.

\bibitem[Villani, 2008]{villani2008optimal}
Villani, C. (2008).
\newblock {\em Optimal Transport: Old and New}, volume 338.
\newblock Springer Science \& Business Media.

\bibitem[Wallner and Dyn, 2005]{wallner2005subdivision}
Wallner, J. and Dyn, N. (2005).
\newblock Convergence and {$C^1$} analysis of subdivision schemes on manifolds
  by proximity.
\newblock {\em Computer Aided Geometric Design}, 22(7):593--622.

\end{thebibliography}
\bibliographystyle{apalike}

\onecolumn 
\appendix
\aistatstitle{Supplementary Material: Efficient Trajectory Inference in Wasserstein Space Using Consecutive Averaging}

\section{SUBDIVISION SCHEMES AND THE LANE RIESENFELD ALGORITHM}\label{app:subdivision-schemes}

\begin{algorithm}[ht]
    \caption{B-spline Approximation Algorithm by \cite{Lane1980LRalgorithm}}
    \label{alg:LRm}
    \begin{algorithmic}[1]
    \Procedure{LaneRiesenfeld}{$[x_j]_{j \in \mathbb{Z}}, R, M$}
    \State \textbf{input} Points to be refined $[x_j]_{j \in \mathbb{Z}}$. $x_j \in \mathbb{R}^d$.
    \State \textbf{input} Refinement level $R \in \mathbb{Z}_{+}$.
    \State \textbf{input} Degree $M$ of B-splines to be approximated.
    \State $y^{(M)} \gets [x_j]_{j\in\mathbb{Z}}$ 
    \Comment{\textit{Initializing points to be doubled.}}
    \For{$r =1$ \textbf{to} $R$ }
    \For{$j \in \mathbb{Z}$} 
        \State $y_{2j}^{(0)} \gets y_j^{(M)}$  \Comment{\textit{Doubling points.}}
        \State $y_{2j+1}^{(0)} \gets y_j^{(M)}$
    \EndFor
    \For{$m = 1$ \textbf{to} $M$} 
        \For{$j \in \mathbb{Z}$}
            \State $y_{j}^{(m)}  \gets \frac{1}{2} (y_{j}^{(m-1)} + y_{j+1}^{(m-1)} )$ 
            \Comment{\textit{Repeated averaging.}}
        \EndFor
    \EndFor
    \EndFor
    \State \Return Refined points $[y_j^{(M)}]_{j \in \mathbb{Z}}$.
    \EndProcedure
\end{algorithmic}
\end{algorithm}

We briefly describe subdivision schemes and their adaptation to the nonlinear domain via nonlinear averaging. Given an ordered sequence of points, $\mathbf{p} =\{p_i\}_{i=0}^T \subset \mathbb{R}^d$, we consider a linear \emph{refinement rule} of the form
\begin{equation}\label{eq:refine}
    S(\mathbf{p})_j = \sum_{i\in \mathbb{Z}} a_{j-2i}p_i, \quad j \in \mathbb{Z},
\end{equation}
where $\{a_i\}_{i\in I} \subset \mathbb{R}$ is the \emph{mask} of the refinement rule $S$. We assume the mask has finitely many non-zero elements, that is, $|I|<\infty$. Note that to avoid any technical boundary treatment, we extend the sequence $\mathbf{p}$ to all of $\mathbb{Z}$ by repeating the starting point $p_0$ and the endpoint $p_T$, i.e., we consider $\ldots, p_0,p_0,p_1,\ldots,p_{T-1},p_T,p_T,\ldots$. 

To obtain finer and finer sequences of $\mathbf{p}$, we consider $S^k(\mathbf{p})$, i.e.\ $S$ is applied $k-$times to $\mathbf{p}$. The repeated application of refinement rules is called a subdivision scheme. A subdivision scheme is called \emph{convergent} if the piecewise linear interpolant to the data $(i2^{-k},S^k(\mathbf{p})_i)$ converges uniformly to a continuous limit curve with values in $\mathbb{R}^d$. See~\cite{cavaretta1989stationary} for more details.

It has been observed in \cite[Theorem 1]{wallner2005subdivision} that any affinely invariant $(\sum_i a_{j-2i}=1$ for all $j$) refinement rule as in (\eqref{eq:refine}) can be (non-uniquely) expressed via the averaging operator $\operatorname{av}$ from \cref{eq:linear-average}. \cite{wallner2005subdivision} further observed that this fact can be used to define refinement rules for sequences $\mathbf{p}$ in Riemannian manifolds by replacing $\operatorname{av}$ by the geodesic average.

\section{DEPENDENCE ON PARAMETERS} \label{sec:runtime}

\begin{figure*}[ht]
    \centering
    \includegraphics[width=0.75\linewidth]{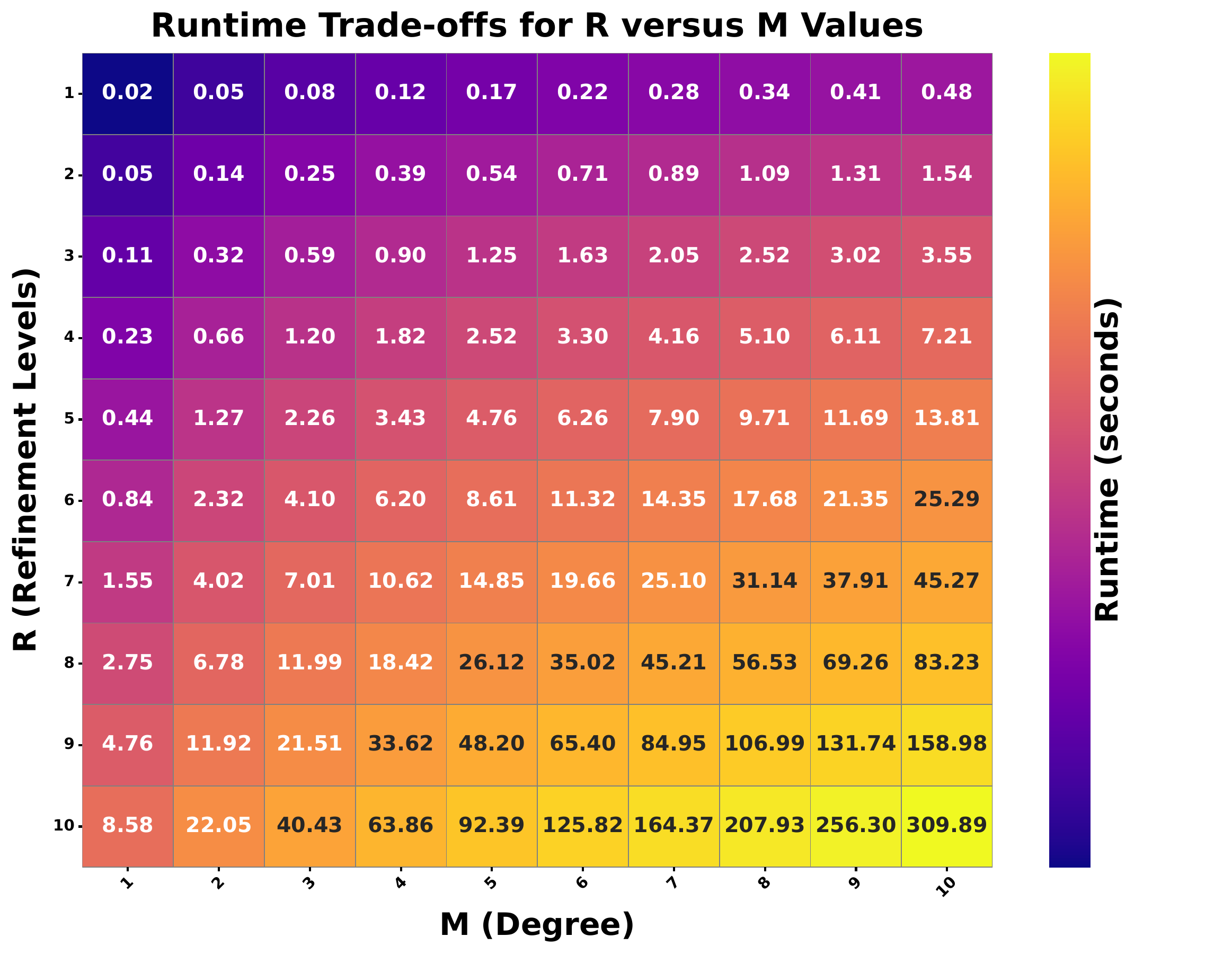}
    \caption{WLR runtimes on the Diverging Gaussian dataset with varying degrees and refinement levels.}
    \label{fig:runtime_tradeoff}
\end{figure*}

The runtime results in Figure \ref{fig:runtime_tradeoff} were obtained by running WLR for varying degrees ($M$) and refinement levels ($R$) on the Diverging Gaussian dataset. It is evident that the runtime increases significantly as $R$ increases, particularly beyond $R=7$. While $M$ also influences runtime, its impact is less pronounced. This is consistent with our analysis in \Cref{sec:complexity} that complexity grows exponentially with $R$ and linearly with $M$. In practice, $R=7,M=3$ is more than enough to achieve smooth trajectories in most cases.

\begin{figure*}[ht]
    \centering
    \begin{subfigure}[b]{0.25\linewidth}
        \includegraphics[width=\linewidth]{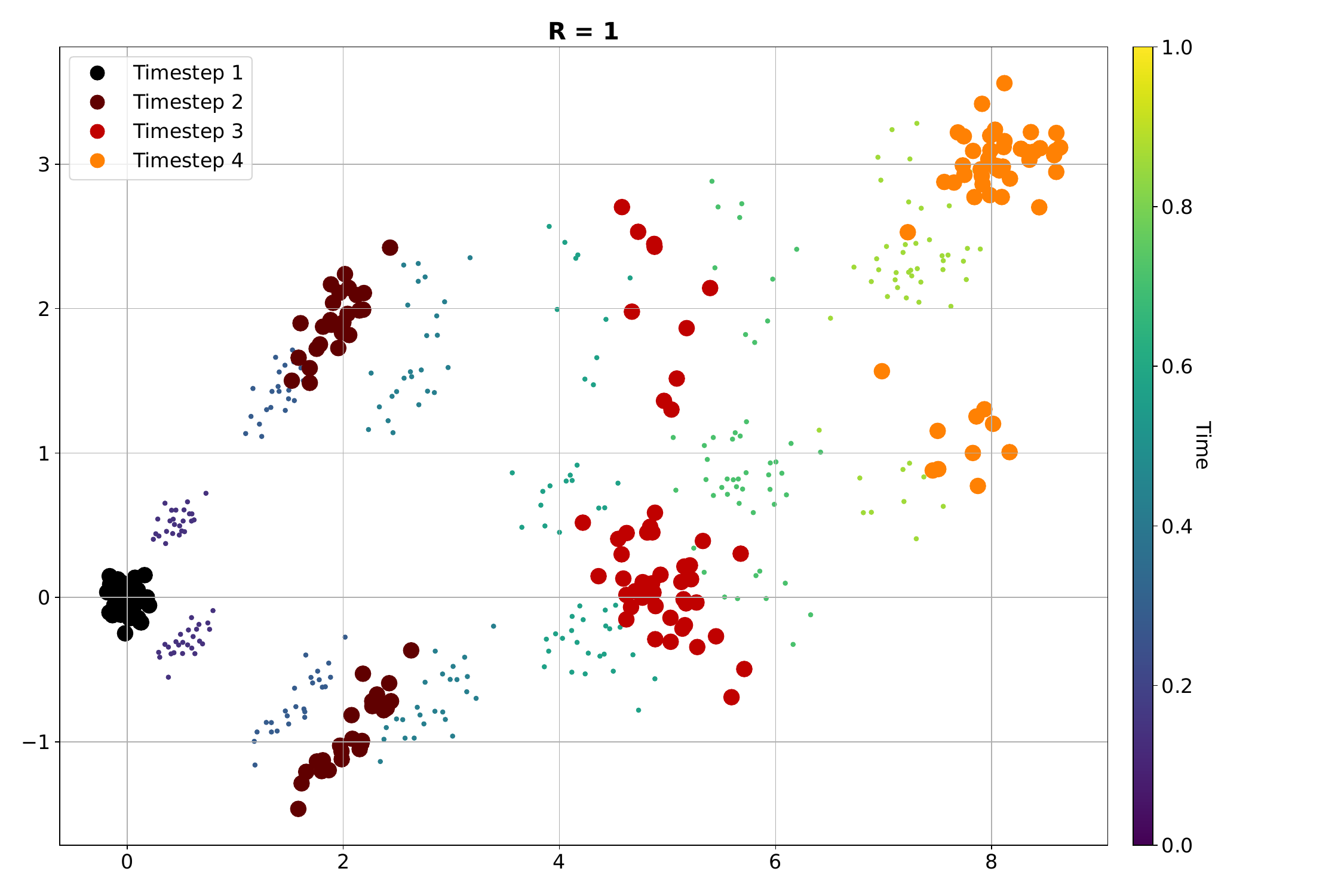}
        \caption{$R=1$}
        \label{fig:wlr_r1}
    \end{subfigure}\hfill
    \begin{subfigure}[b]{0.25\linewidth}
        \includegraphics[width=\linewidth]{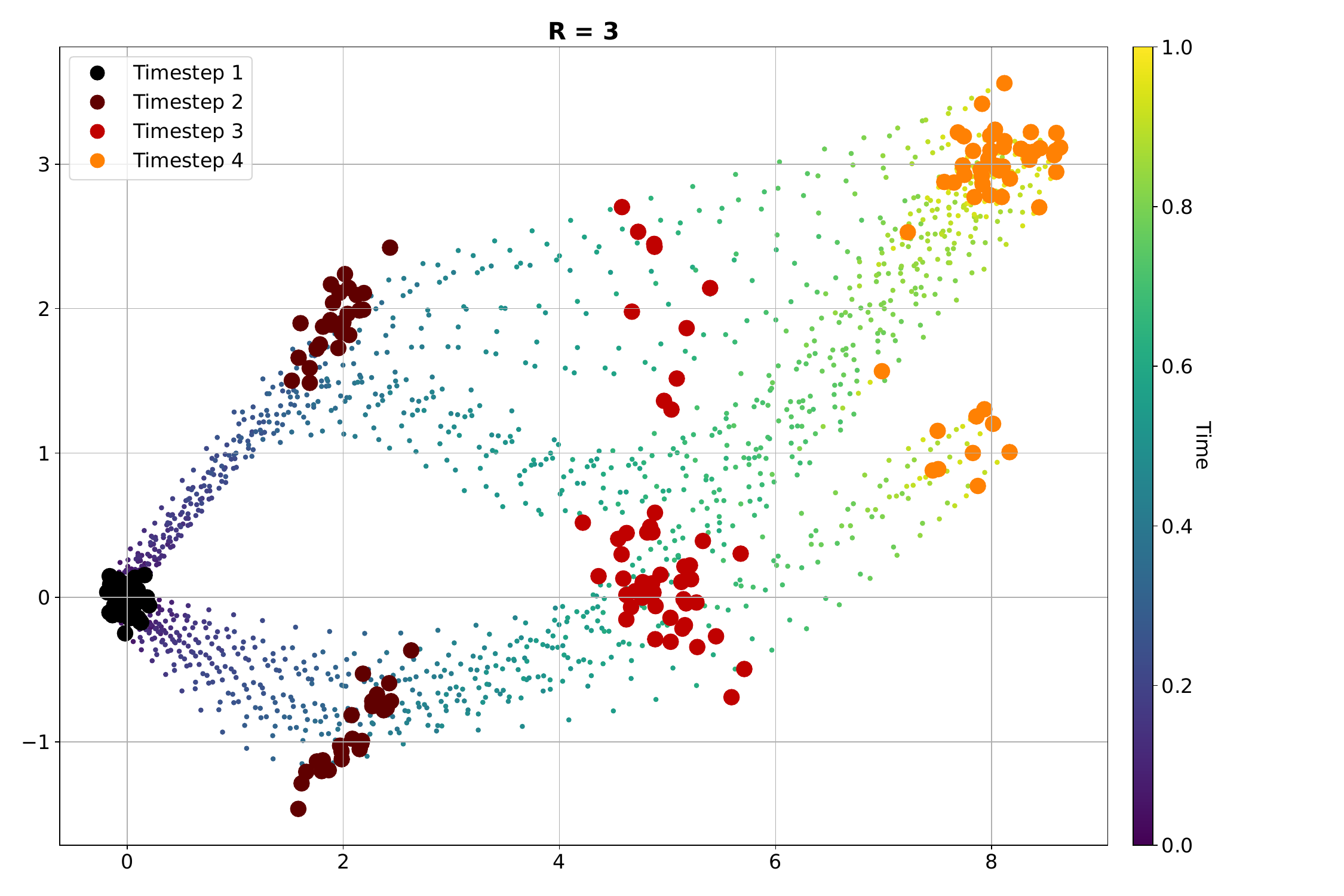}
        \caption{$R=3$}
        \label{fig:wlr_r3}
    \end{subfigure}\hfill
    \begin{subfigure}[b]{0.25\linewidth}
        \includegraphics[width=\linewidth]{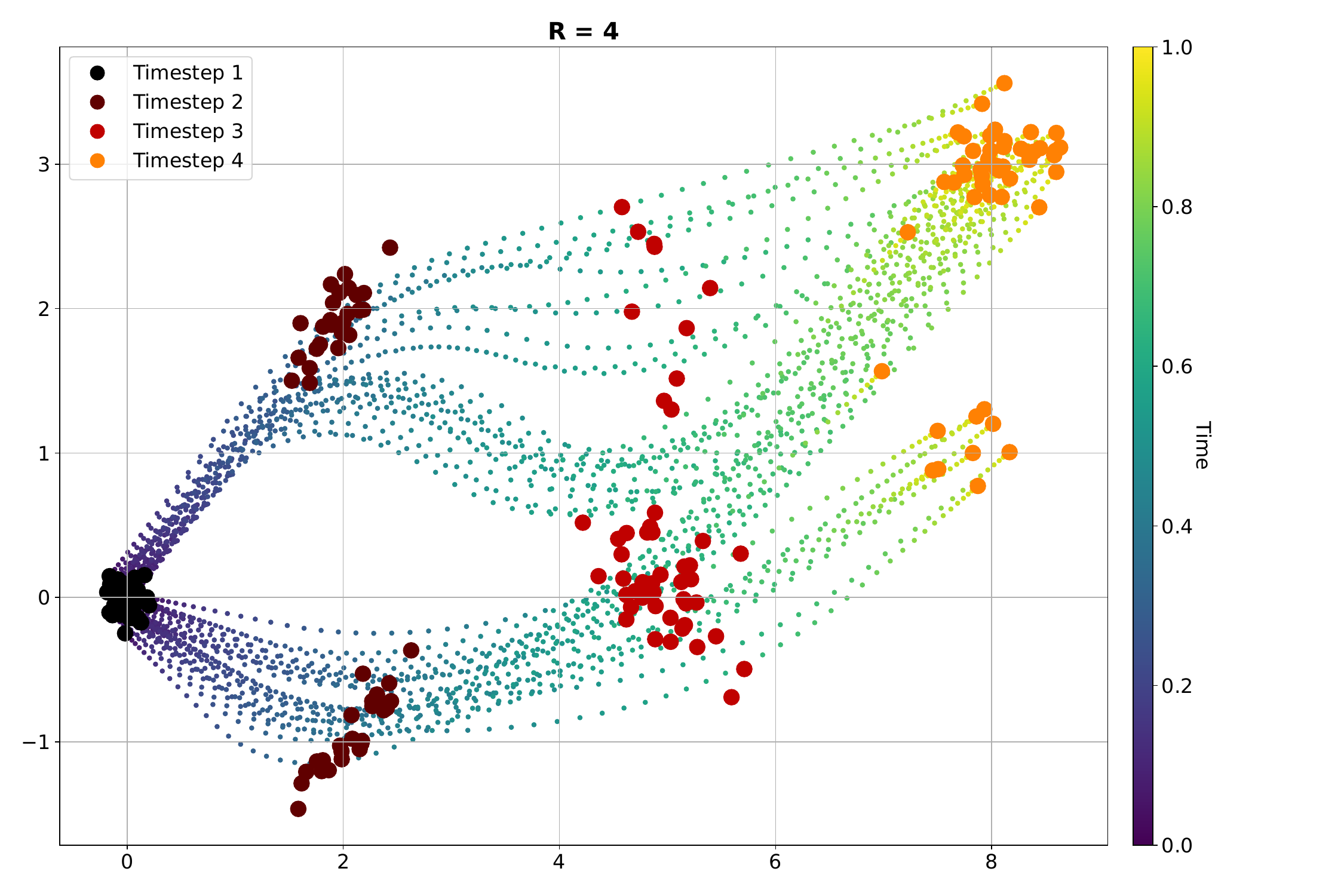}
        \caption{$R=4$}
        \label{fig:wlr_r4}
    \end{subfigure}\hfill
    \begin{subfigure}[b]{0.25\linewidth}
        \includegraphics[width=\linewidth]{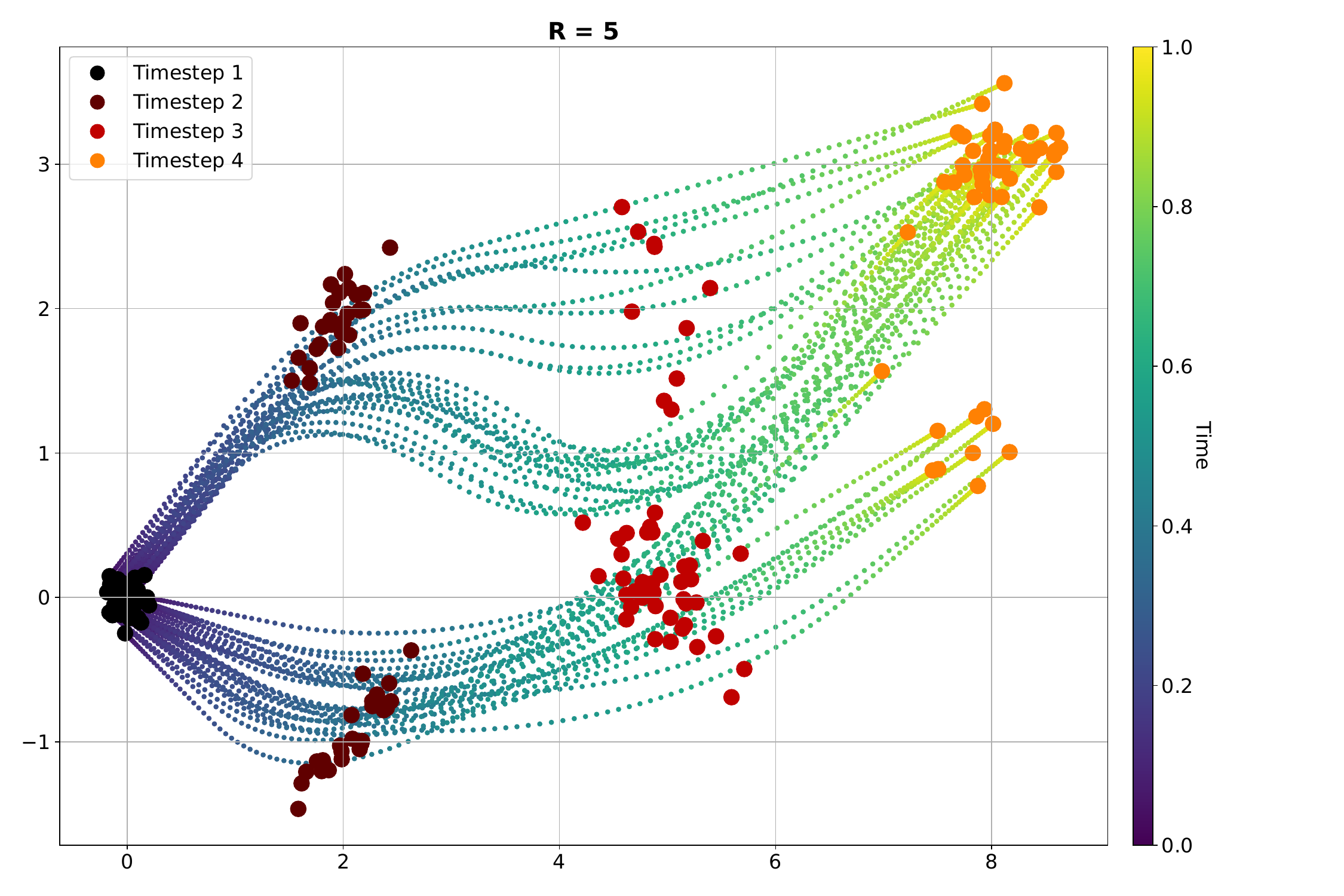}
        \caption{$R=5$}
        \label{fig:wlr_r5}
    \end{subfigure}
    \caption{WLR on Diverging Gaussians at various values of $R$.}
    \label{fig:wlr-diverging-gaussians-eg}
\end{figure*}

\Cref{fig:wlr-diverging-gaussians-eg} shows the application of WLR on the Diverging Gaussian data at various refinement levels, with lower refinement levels offering reduced computational time. This highlights the algorithm’s adaptability to user-defined refinement levels. In our implementation, the number of refined points returned by WLR also depends on $R$ and $M$ as $|\nu^{(M)}| = 2^R (T+M  -1) + 2-M$, which follows from Algorithm~\ref{alg:WLRm}.

\section{PRELIMINARY RESULTS: WLR AS INITIALIZATION FOR DEEP LEARNING-BASED METHODS}  \label{sec:NN_initialization}

\begin{figure}[H]
    \centering
    \begin{subfigure}[b]{0.48\linewidth}
        \centering
        \includegraphics[width=\linewidth]{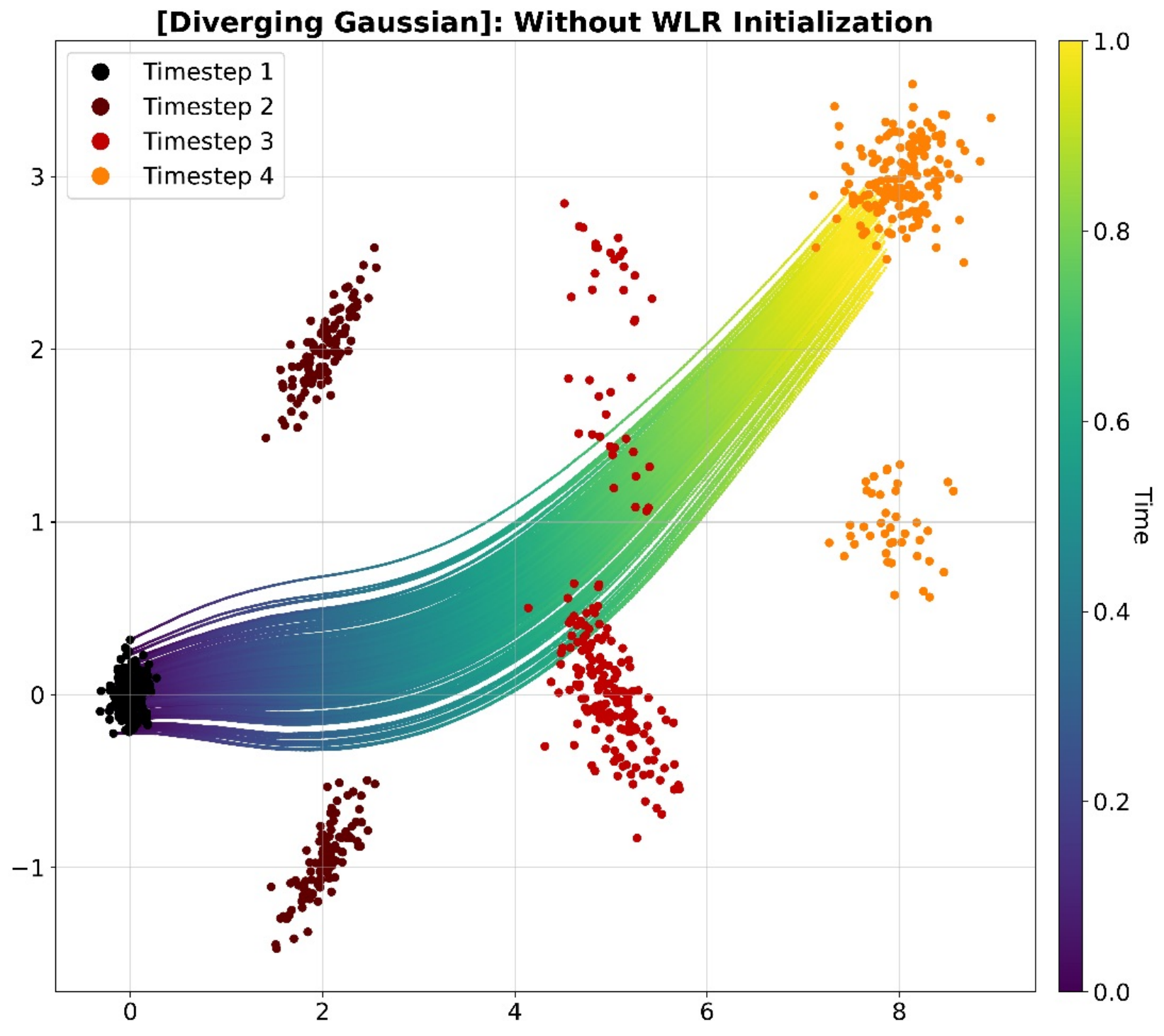}
        \caption{MIOFlow}
        \label{fig:sub-figure-1-WLRinit}
    \end{subfigure}
    \hfill
    \begin{subfigure}[b]{0.48\linewidth}
        \centering
        \includegraphics[width=\linewidth]{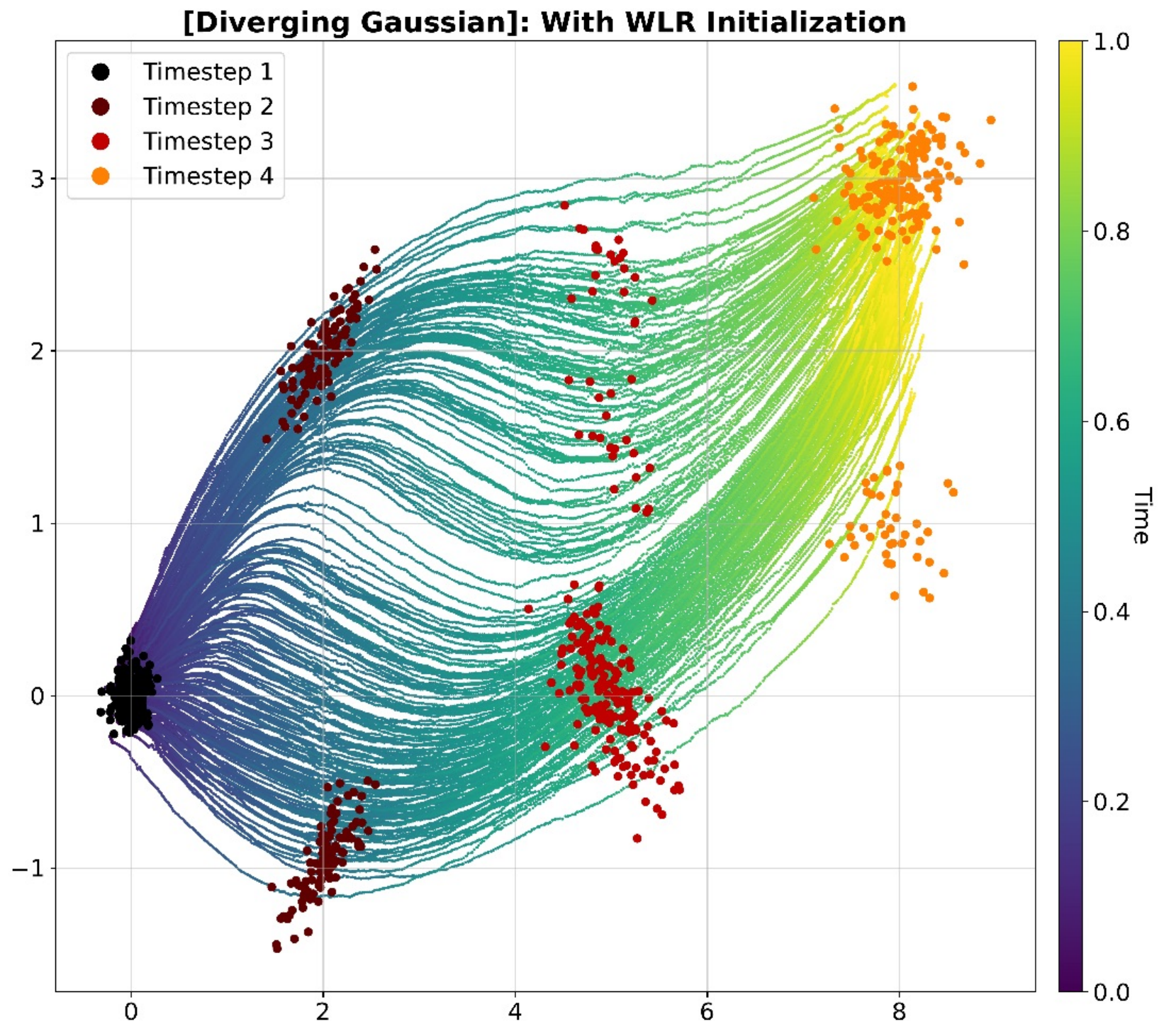}
        \caption{WLR + MIOFlow}
        \label{fig:sub-figure-2-WLRinit}
    \end{subfigure}
    \vskip\baselineskip
    \caption{On the left, \Cref{fig:sub-figure-1-WLRinit} shows MIOFlow trajectories without any prior initialization. \Cref{fig:sub-figure-2-WLRinit} shows the trajectories after initialization with WLR on Diverging Gaussian Data. In both cases MIOFlow was run with a fixed budget of 10 epochs}
    \label{fig:main-figure-WLRinit}
\end{figure}

While a comprehensive study of WLR initialization for deep learning methods remains a promising direction for future work, our preliminary experiments suggest some qualitative benefits. To demonstrate these advantages, we compare two scenarios: training MIOFlow from scratch and training MIOFlow initialized with WLR trajectories, both with a fixed computational budget of 10 epochs.
The qualitative results can be seen in \Cref{fig:main-figure-WLRinit}. MIOFlow alone struggles to capture the bifurcating behavior present in the diverging Gaussian dataset, producing simplified trajectories without branching as seen in (\Cref{fig:sub-figure-1-WLRinit}). In contrast, when initialized with WLR trajectories, MIOFlow successfully maintains the bifurcating structure while benefiting from the neural network's capacity to refine the trajectories (\Cref{fig:sub-figure-2-WLRinit}). This result hints that WLR initialization could provide geometric guidance that helps in capturing complex trajectories.

\section{EXPERIMENTAL DETAILS} 
\label{app:experiment}
We provide details on the experiments in \Cref{sec:experiments}. 

\subsection{Datasets} \label{sec:datasets}
We use a wide range of simulated and real datasets that capture the natural dynamics observed in cellular differentiation. This includes phenomena such as bifurcations and merges, as depicted in Figure \ref{fig:WLR_interp}.

\paragraph{Diverging Gaussian}
We designed this dataset (\Cref{fig:WLR_interp}a) to initiate from a singular point cluster and gradually disperse as time progresses. The points all have uniform mass.

\paragraph{Petal}
The Petal dataset (\Cref{fig:WLR_interp}b), as defined by \cite{huguet2022manifold}, presents a significant challenge due to its inherent geometric structure. We resampled it to make $n_t$ consistent. The points all have uniform mass.

\paragraph{Dyngen Tree and Cycle}
Utilizing Dyngen by \cite{dyngen_cannoodt2021}, we simulated a single-cell RNA sequencing (scRNA-seq) dataset to model a dynamic cellular process. We generated both tree (\Cref{fig:WLR_interp}d) and cycle (\Cref{fig:WLR_interp}e) data, using 100 transcription factors for each. They were then embedded into a 10-dimensional space via PCA and PHATE (\cite{phate_moon2019visualizing}), respectively.

\paragraph{CITE-seq} 
This dataset from the Multimodal Single-Cell Integration challenge at NeurIPS 2022 \cite{citeseq_stoeckius2017large} consists of time-series measurements of CD34+ hematopoietic stem and progenitor cells (HSPCs) collected at four distinct time points: days 2, 3, 4, and 7 post-isolation. To prepare the data for trajectory inference, we normalized each feature dimension to ensure comparable scales across features. We then computed the first 100 principal components of the data to capture the most salient features of cellular variation and to maintain a higher dimensionality compared to other datasets.

\paragraph{Converging Gaussian}
This dataset (\Cref{fig:WLR_interp}c) was crafted by \cite{clancy2022wassersteinfisherrao} to simulate the intricate dynamics characteristic of cellular differentiation. The points are sampled from Gaussian distributions with intermediate stages having cellular division processes. Changing $n_t$ across intermediate time steps makes this dataset challenging. F\&S cannot handle this setup, so we skip it. On the other hand, we were not able to adapt code by \cite{clancy2022wassersteinfisherrao} to run on any other datasets, so we report results from Wasserstein-Fisher-Rao only on Converging Gaussian. We also do not calculate MSE since the definition becomes ambiguous as the number of points between the predicted point clouds and the held out point clouds can differ.

\paragraph{CITE-seq Supercells}

We used \cite{supercell_bilous2022metacells}\footnote{\url{https://github.com/GfellerLab/SuperCell}} on the full CITE-seq dataset to create supercell (metacells), which are aggregates of individual cells that share similar transcriptomic profiles. SuperCells package constructs a $k$-nearest neighbor ($k$-NN) graph based on the most variable genes, then partitions this graph to form metacells. Gene expression profiles for these metacells are computed by averaging the expression levels of their constituent cells. We used a graining level of $\gamma = 20$ with $k=5$.

\subsection{Parameters and Setup}

All experiments including runtime calculations were carried out on x86\_64 machine with AMD EPYC 7713 64-Core Processor with NVIDIA A100-PCIE-40GB GPU. WLR, F\&S and Wasserstein-Fisher-Rao did not require any GPU for training. Our paper used two software packages: Python OT\footnote{\url{https://pythonot.github.io/}} version 0.9.3 with MIT license from \cite{PythonOT} and Dyngen\footnote{\url{https://github.com/dynverse/dyngen}} version 1.0.5 with MIT license (CC-BY) from \cite{dyngen_cannoodt2021}. Dataset details are in \Cref{tab:dataset_details}. For the experiments in \Cref{tab:experiments}, data subsampling was carried out using the following seeds: 1783345717, 3621289926, 3137468471, 1383545203, 3849164755.

\begin{table*}[htp]
    \centering
    \footnotesize
    \begin{tabular}{lcccccc} \toprule
    \textbf{Dataset} &  $d$ & $T+1$ & $j$& Original \# of points &  $n_t$ & \textbf{Author}\\ \midrule \addlinespace[1mm] 
    Diverging Gaussian    & 2  & 4 & $2$ & 200 & -           & Us \\
    Petal                & 2  & 5  & $2$ & 135 & -           & \cite{huguet2022manifold} \\
    Converging Gaussian   & 2  & 4 & $2$ & 126, 378, 252, 126  & - & \cite{clancy2022wassersteinfisherrao} \\
    Dyngen Tree           & 10 & 7 & $ 3$ & 319, 665, 992, 993, 743, 460, 328 & 319           & \cite{dyngen_cannoodt2021} \\
    Dyngen Cycle          & 10 & 15 & $8$ & 52 & -            & \cite{dyngen_cannoodt2021} \\
    CITE-seq               & 100 & 4 & $2$ & 6071, 7643, 8485, 7195 & 1000          & \cite{citeseq_stoeckius2017large} \\
    CITE-seq Supercells    & 100 & 4 & $2$ & 304, 382, 424, 360 & -          & Us \\ \bottomrule
    \end{tabular}
    \caption{Dimensions $d$, number of time steps $T+1$, held out time step $j$, original number of points, and uniformly sampled (if applicable) number of points $n_t$ of the datasets used in \Cref{sec:experiments}.}
    \label{tab:dataset_details}
\end{table*}

\paragraph{Wasserstein Lane-Riesenfeld (WLR)}
We used $M=10$ for Petal and $M=2$ for the rest. $R = 7$ and $\epsilon = 10^{-5}$ in all experiments in \Cref{sec:sota-comparison}. For our experiments on supercells, we used $R = 4, 5, 6$ and $\epsilon = 10^{-8}$. 

\paragraph{TrajectoryNet}
 For all experiments, we used TrajectoryNet\footnote{\url{https://github.com/KrishnaswamyLab/TrajectoryNet}} with 1000 iterations and \texttt{whiten=True} with all other default optimization parameters in \cite{tong2020trajectorynet}. 
 
\paragraph{MIOFLow}
We observed the best trajectory inference results without training the geodesic autoencoder but utilizing the density loss for all of our experiments. We used the following hyperparameters for MIOFlow\footnote{\url{https://github.com/KrishnaswamyLab/MIOFlow}}. Despite our best efforts, we were unable to reproduce their published figure on Petal. After multiple attempts, we report their \textit{best} outcome in \Cref{tab:experiments} and \Cref{fig:full_scale_experiemnts}.
\begin{itemize}
    \item Diverging Gaussian: sample\_size=60, n\_local\_epochs = 50, n\_epochs = 0, n\_post\_local\_epochs = 0, $\lambda_e = 0.001$, $\lambda_d = 10$, $\epsilon=0.5$ n\_points = 200, n\_trajectories = 200, n\_bins = 500, n\_epochs\_emb = 1000
    \item Converging Gaussian: sample\_size=30, n\_local\_epochs = 50, n\_epochs = 0, n\_post\_local\_epochs = 0, $\lambda_e = 0.001$, $\lambda_d = 35$, $\epsilon=0.5$ n\_points = 64, n\_trajectories = 64, n\_bins = 500, n\_epochs\_emb = 1000
    \item Petal: n\_local\_epochs = 50, n\_points = 135, n\_trajectories = 135, n\_bins = 500. The rest are default hyperparameters mentioned in \cite{huguet2022manifold}.
    \item Dyngen Tree: sample\_size=30, n\_local\_epochs = 5, n\_epochs = 0, n\_post\_local\_epochs = 0, $\lambda_e = 0.001$, $\lambda_d = 15$, $\epsilon=0.5$ n\_points = 100, n\_trajectories = 100, n\_bins = 500, n\_epochs\_emb = 1000
    \item Dyngen Cycle: sample\_size=30, n\_local\_epochs = 10, n\_epochs = 0, n\_post\_local\_epochs = 0, $\lambda_e = 0.001$, $\lambda_d = 15$, $\epsilon=0.5$ n\_points = 50, n\_trajectories = 50, n\_bins = 500, n\_epochs\_emb = 1000
    \item CITE-seq: sample\_size=100, n\_local\_epochs = 50, n\_epochs = 0, n\_post\_local\_epochs = 0, $\lambda_e = 0.001$, $\lambda_d = 15$, $\epsilon=0.5$ n\_points = 50, n\_trajectories = 1000, n\_bins = 500, n\_epochs\_emb = 1000, use\_density\_loss = True
\end{itemize}

\paragraph{Fast and Smooth (F \& S)}
The original authors of \cite{chewi2021fast} proposed a general algorithm for performing interpolation using cubic splines in $\mathbb{R}^d$ after determining an ordering through successive optimal transport computations on the initially observed point clouds. To ensure a fair comparison, we wrote our own code to adapt their methodology to approximate B-splines in $\mathbb{R}^d$ via the Lane-Riesenfeld algorithm {and carried out comparisons with both the versions (B-splines and cubic splines) of this algorithm. We used the same $M$ and $R$ as WLR. As mentioned in \Cref{sec:datasets}, F\&S is not able to run on Converging Gaussian.

\paragraph{Wasserstein-Fisher-Rao}
The authors of \cite{clancy2022wassersteinfisherrao} proposed a method for computing splines for measures of differing masses using the notion of unbalanced optimal transport. We used their method for computing splines on the converging Gaussian data using all default parameters with $\eta_{\text{OT}} = 0.001$ for 200 iterations and 32 interpolating splines from the first to the last time step.  
As mentioned in \Cref{sec:datasets}, we were able to run their code\footnote{\url{https://github.com/felipesua/WFR_splines}} only on this dataset.

\paragraph{Conditional Flow Matching}
We used batch\_size = 256, $\sigma = 0.1$, $w=64$ and learning\_rate = 0.0001, epochs = 10000 and n\_traj\_bins = 500 for all our experiments. The code can be found here \footnote{\url{https://github.com/atong01/conditional-flow-matching}}.
\begin{itemize}
    \item Model Architecture:
    \begin{itemize}
        \item Network: Multi-Layer Perceptron (MLP) with default layers and size and activations
        \item Input Dimension: Concatenation of position coordinates (data dimensions) and time
        \item Time-Varying: Yes
        \item Output Dimension: Matching data dimensionality
    \end{itemize}
    
    \item Training Parameters:
    \begin{itemize}
        \item Batch size: 256
        \item Number of iterations: 10,000
        \item Learning rate: 1e-4 (Adam optimizer)
        \item Sigma ($\sigma$): 0.1
        \item Weight Decay: 0 (default)
    \end{itemize}
    
    \item Loss Function:
    \begin{itemize}
        \item Type: Mean Squared Error (MSE) between predicted and true conditional flows
        \item Additional Terms: None
    \end{itemize}
    
    \item Trajectory Generation:
    \begin{itemize}
        \item Number of Trajectory Bins: 500
        \item Trajectory Sampling: 
        \begin{itemize}
            \item ODE Solver: Dormand-Prince (dopri5)
            \item Sensitivity Method: Adjoint
        \end{itemize}
    \end{itemize}
\end{itemize}

\paragraph{Wasserstein Lagrangian Flow (WLF)}
We implemented the algorithm proposed by \cite{wlflow_neklyudov_computational_2023} and compared their method using two loss variants, namely unbalanced optimal transport with physical potential (UBOT+) and Schr\"{o}dinger Bridge (SB). For all experiments, we used the following hyperparameters:

\begin{itemize}
    \item Model Architecture:
    \begin{itemize}
        \item Score Network (S): 4-layer MLP with 512 hidden units and Swish activation
        \item Sampling Network (Q): 4-layer MLP with 128 hidden units and Swish activation
        \item Both networks use skip connections and dropout rate of 0.2
    \end{itemize}
    
    \item Training Parameters:
    \begin{itemize}
        \item Batch size: 512
        \item Number of iterations: 1000
        \item Learning rate: 1e-3 (Adam optimizer)
        \item Step size: 0.05
    \end{itemize}
    
    \item Loss-Specific Parameters:
    \begin{itemize}
        \item SB variant: noise level $\sigma = 0.1$
        \item UBOT+ variant: balance parameter $\lambda = 1.0$
    \end{itemize}
    
    \item Sampling:
    \begin{itemize}
        \item Latent dimension for sampling network: 10
        \item Number of simulation steps: 500
    \end{itemize}
\end{itemize}

 \section{ADDITIONAL RESULTS} 
\label{app:additional_results}

We present full visual comparison between WLR and other methods in \Cref{fig:full_scale_experiemnts}. Trajectory inference is done on all time steps with no hold-out point clouds. \Cref{tab:experiments} contains experimental results on datasets having point clouds with uniform mass.

\begin{figure*}[htp]
    \centering
    \includegraphics[width=\linewidth]{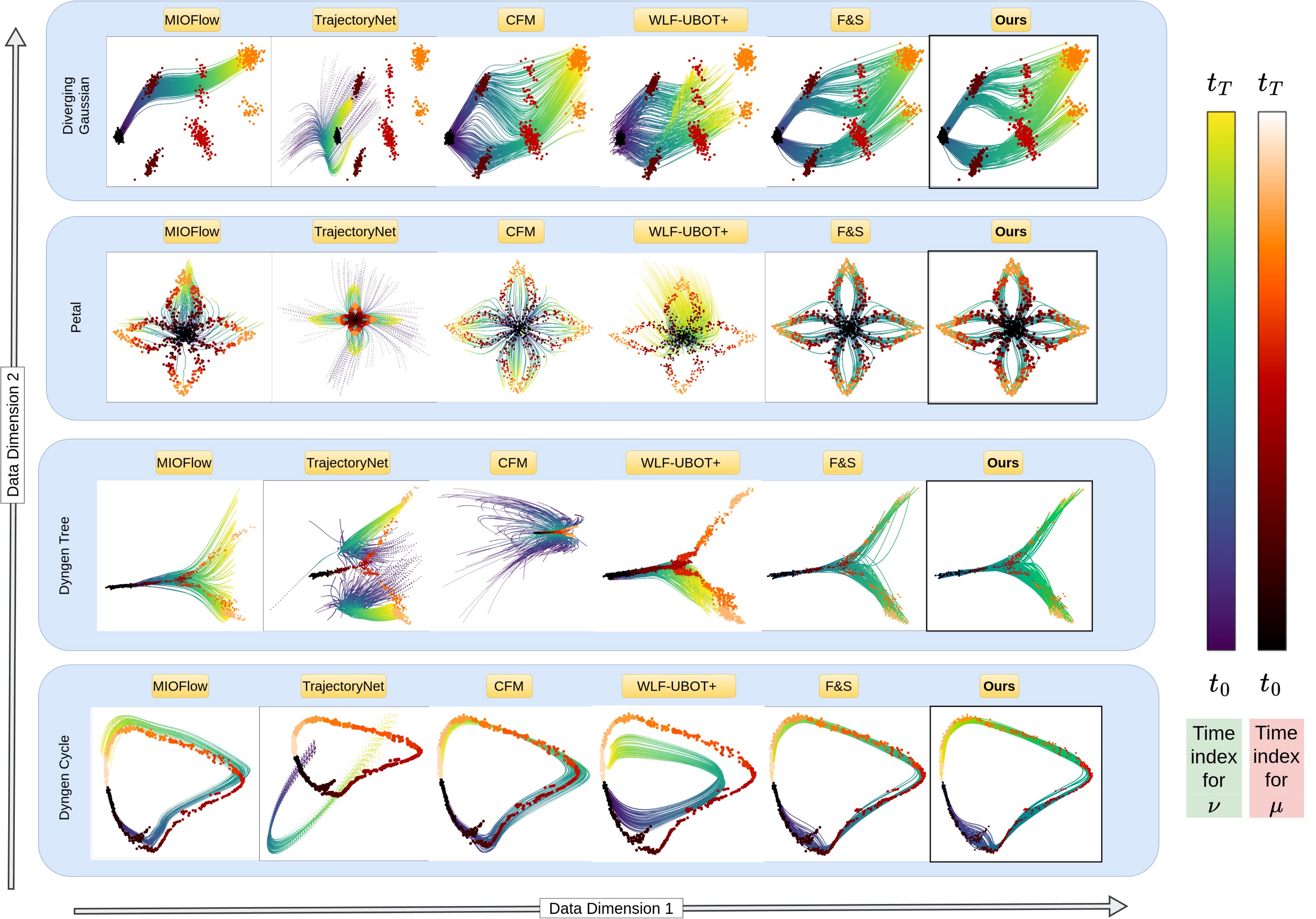}
    \caption{Full-scale visual comparison between WLR and other methods with point clouds having uniform mass and equal points per time step. Trajectory inference is done on all time steps with no hold-out point clouds. Note that we used F \& S with B-splines to ensure a fair comparison with WLR, as both are approximation schemes.}
    \label{fig:full_scale_experiemnts}
\end{figure*}

\begin{table*}[htp]
\centering
\setlength{\tabcolsep}{4pt}
\scriptsize
\begin{tabular}{l|l|c|cc|cc}
\toprule
\multirow{2}{*}{Dataset} & \multirow{2}{*}{Method} & Runtime & \multicolumn{2}{|c|}{Leave-one-out} & \multicolumn{2}{|c}{Mean} \\
& & (sec) $\downarrow$ & MSE $\downarrow$ & $W_1 \downarrow$ & MSE $\downarrow$ & $W_1 \downarrow$ \\
\midrule
\multirow{7}{*}{Diverging Gaussian}
& WLR (m$=$2)  (Ours)     
  & $\underline{3.99 \pm 0.03}$
  & $\mathbf{1.73 \pm 0.06}$
  & $\mathbf{1.16 \pm 0.02}$
  & $\underline{1.16 \pm 0.81}$
  & $\mathbf{0.61 \pm 0.61}$ \\
& CFM              
  & $182.07 \pm 0.67$
  & $\underline{1.92 \pm 0.06}$
  & $1.45 \pm 0.02$
  & $\mathbf{1.12 \pm 0.81}$
  & $\underline{0.66 \pm 0.57}$ \\
& MIOFlow          
  & $16.85 \pm 0.16$
  & $2.18 \pm 1.69$
  & $1.58 \pm 0.76$
  & $1.61 \pm 1.15$
  & $1.19 \pm 0.76$ \\
& TrajectoryNet    
  & $1340.30 \pm 88.83$
  & $14.04 \pm 0.21$
  & $5.30 \pm 0.04$
  & $12.02 \pm 9.42$
  & $4.37 \pm 2.11$ \\
& F \& S (B Spline)
  & $\mathbf{1.14 \pm 0.01}$
  & $\mathbf{1.73 \pm 0.06}$
  & $\mathbf{1.16 \pm 0.02}$
  & $\underline{1.16 \pm 0.81}$
  & $\mathbf{0.61 \pm 0.61}$ \\
& WLF-SB           
  & $103.08 \pm 10.52$
  & $2758.34 \pm 5987.36$
  & $41.16 \pm 68.71$
  & $2799.99 \pm 6063.34$
  & $41.98 \pm 68.79$ \\
& WLF-UBOT+        
  & $89.71 \pm 12.08$
  & $1.96 \pm 0.33$
  & $\underline{1.16 \pm 0.18}$
  & $7.93 \pm 0.43$
  & $3.29 \pm 0.05$ \\
\midrule
\multirow{7}{*}{Petal}
& WLR (m$=$2)  (Ours)     
  & $\mathbf{25.27 \pm 0.97}$
  & $0.26 \pm 0.01$
  & $\mathbf{0.09 \pm 0.01}$
  & $\mathbf{0.01 \pm 0.01}$
  & $\mathbf{0.05 \pm 0.05}$ \\
& CFM              
  & $233.26 \pm 1.10$
  & $\mathbf{0.23 \pm 0.01}$
  & $0.18 \pm 0.02$
  & $0.30 \pm 0.27$
  & $0.14 \pm 0.08$ \\
& MIOFlow          
  & $45.15 \pm 0.19$
  & $0.26 \pm 0.02$
  & $0.26 \pm 0.09$
  & $0.32 \pm 0.27$
  & $0.24 \pm 0.16$ \\
& TrajectoryNet    
  & $1415.57 \pm 56.44$
  & $0.49 \pm 0.02$
  & $0.31 \pm 0.01$
  & $0.81 \pm 0.94$
  & $0.64 \pm 0.40$ \\
& F \& S (B Spline)
  & $36.59 \pm 0.42$
  & $0.27 \pm 0.01$
  & $\underline{0.11 \pm 0.01}$
  & $0.36 \pm 0.33$
  & $\underline{0.11 \pm 0.10}$ \\
& WLF-SB           
  & $53.44 \pm 6.01$
  & $1631.28 \pm 2818.99$
  & $38.75 \pm 46.76$
  & $1630.06 \pm 2816.59$
  & $38.74 \pm 46.74$ \\
& WLF-UBOT+        
  & $\underline{35.79 \pm 4.26}$
  & $\underline{0.25 \pm 0.15}$
  & $0.42 \pm 0.26$
  & $\underline{0.28 \pm 0.15}$
  & $0.47 \pm 0.24$ \\
\midrule
\multirow{7}{*}{Dyngen Tree}
& WLR (m$=$2)  (Ours)     
  & $13.18 \pm 0.09$
  & $\mathbf{0.82 \pm 0.02}$
  & $\mathbf{1.76 \pm 0.01}$
  & $\mathbf{1.55 \pm 1.13}$
  & $\mathbf{1.32 \pm 0.94}$ \\
& CFM              
  & $423.42 \pm 0.38$
  & $2.02 \pm 0.40$
  & $3.31 \pm 0.39$
  & $17.68 \pm 28.68$
  & $6.69 \pm 7.26$ \\
& MIOFlow          
  & $\underline{6.98 \pm 0.12}$
  & $2.17 \pm 0.90$
  & $3.75 \pm 0.91$
  & $16.74 \pm 26.93$
  & $6.34 \pm 6.16$ \\
& TrajectoryNet    
  & $2692.71 \pm 139.31$
  & $\underline{1.38 \pm 0.20}$
  & $\underline{2.97 \pm 0.32}$
  & $\underline{1.91 \pm 1.33}$
  & $\underline{3.14 \pm 1.09}$ \\
& F \& S (B Spline)
  & $\mathbf{3.65 \pm 0.02}$
  & $\mathbf{0.82 \pm 0.02}$
  & $\mathbf{1.76 \pm 0.01}$
  & $\mathbf{1.55 \pm 1.13}$
  & $\mathbf{1.32 \pm 0.94}$ \\
& WLF-SB           
  & $52.68 \pm 7.63$
  & $10.27 \pm 5.10$
  & $9.59 \pm 2.41$
  & $9.99 \pm 4.98$
  & $9.26 \pm 2.26$ \\
& WLF-UBOT+        
  & $39.32 \pm 7.79$
  & $2.53 \pm 0.37$
  & $4.34 \pm 0.44$
  & $3.13 \pm 0.44$
  & $4.52 \pm 0.49$ \\
\midrule
\multirow{7}{*}{Dyngen Cycle}
& WLR (m$=$2)  (Ours)     
  & $\underline{1.62 \pm 0.01}$
  & $\mathbf{0.06 \pm 0.01}$
  & $\mathbf{0.58 \pm 0.01}$
  & $\mathbf{0.09 \pm 0.04}$
  & $\mathbf{0.73 \pm 0.32}$ \\
& CFM              
  & $1055.50 \pm 2.92$
  & $0.11 \pm 0.05$
  & $0.77 \pm 0.21$
  & $0.76 \pm 1.29$
  & $0.85 \pm 0.54$ \\
& MIOFlow          
  & $26.86 \pm 0.12$
  & $\mathbf{0.06 \pm 0.02}$
  & $\underline{0.65 \pm 0.12}$
  & $0.11 \pm 0.07$
  & $\underline{0.75 \pm 0.37}$ \\
& TrajectoryNet    
  & $5758.90 \pm 261.24$
  & $14.50 \pm 2.55$
  & $11.99 \pm 1.04$
  & $7.08 \pm 5.36$
  & $7.58 \pm 3.51$ \\
& F \& S (B Spline)
  & $\mathbf{1.36 \pm 0.01}$
  & $\mathbf{0.06 \pm 0.01}$
  & $\mathbf{0.58 \pm 0.01}$
  & $\underline{0.10 \pm 0.04}$
  & $\mathbf{0.73 \pm 0.32}$ \\
& WLF-SB           
  & $93.09 \pm 10.48$
  & $222.18 \pm 306.67$
  & $37.46 \pm 31.30$
  & $218.33 \pm 304.44$
  & $36.60 \pm 31.75$ \\
& WLF-UBOT+        
  & $79.23 \pm 9.78$
  & $2.46 \pm 0.38$
  & $4.92 \pm 0.36$
  & $1.51 \pm 0.32$
  & $3.60 \pm 0.37$ \\
\midrule
\multirow{7}{*}{CITE-seq}
& WLR (m$=$2)  (Ours)     
  & $120.91 \pm 0.43$
  & $35.54 \pm 0.48$
  & $\mathbf{31.72 \pm 0.15}$
  & $\underline{31.70 \pm 20.70}$
  & $\mathbf{12.14 \pm 13.23}$ \\
& CFM              
  & $186.30 \pm 2.01$
  & $37.36 \pm 0.70$
  & $34.32 \pm 0.33$
  & $34.27 \pm 23.25$
  & $\underline{24.92 \pm 14.84}$ \\
& MIOFlow          
  & $\underline{90.54 \pm 1.21}$
  & $31.27 \pm 0.84$
  & $32.43 \pm 0.40$
  & $34.50 \pm 8.79$
  & $27.08 \pm 10.87$ \\
& TrajectoryNet    
  & $10349.98 \pm 192.70$
  & $\mathbf{11.21 \pm 0.14}$
  & $\underline{31.75 \pm 0.23}$
  & $\mathbf{20.06 \pm 7.96}$
  & $39.84 \pm 7.03$ \\
& F \& S (B Spline)
  & $\mathbf{6.41 \pm 0.03}$
  & $35.54 \pm 0.48$
  & $\mathbf{31.72 \pm 0.15}$
  & $\underline{31.70 \pm 20.70}$
  & $\mathbf{12.14 \pm 13.23}$ \\
& WLF-SB           
  & $98.58 \pm 5.63$
  & $596.42 \pm 489.34$
  & $220.59 \pm 93.82$
  & $602.36 \pm 488.36$
  & $220.97 \pm 93.50$ \\
& WLF-UBOT+        
  & $105.10 \pm 8.29$
  & $\underline{30.15 \pm 0.28}$
  & $32.58 \pm 0.20$
  & $34.13 \pm 0.26$
  & $29.65 \pm 0.12$ \\
\bottomrule
\end{tabular}
\caption{Quantitative comparison of approximation methods (non-interpolatory) for trajectory inference. Best scores are in \textbf{bold} and second best are \underline{underlined}. F \& S (B Spline) is our modified version of F \& S proposed by \cite{chewi2021fast}. The methods in this table include B-spline based approximation methods and neural network approaches, both of which provide smooth approximations rather than exact interpolations of the input point clouds. While our approach and F\&S often appear numerically similar, this is expected: the original F\&S method in \cite{chewi2021fast} was limited to cubic spline interpolation in $\mathbb{R}^d$, so we implemented a B-spline approximation version of F\&S (via Lane–Riesenfeld), matching our own B-spline setting and thereby yielding closely aligned scores. Importantly, our method can naturally handle mass splitting and varying numbers of points across time, which F\&S and many other interpolation methods cannot address. Hence, even though our numerical metrics coincide with F\&S under comparable spline degrees, our approach offers greater flexibility in modeling realistic biological processes, such as cell differentiation with bifurcations or supercell formation, where the total mass and number of points vary over time.}
\label{tab:experiments}
\end{table*}

\begin{table*}[htp]
\centering
\setlength{\tabcolsep}{4pt} 
\scriptsize
\begin{tabular}{l|l|c|cc|cc}
\toprule
\multirow{2}{*}{Dataset} & \multirow{2}{*}{Method} & Runtime & \multicolumn{2}{c|}{Leave-one-out} & \multicolumn{2}{c}{Mean} \\
& & (sec) $\downarrow$ & MSE $\downarrow$ & $W_1 \downarrow$ & MSE $\downarrow$ & $W_1 \downarrow$ \\
\midrule
\multirow{2}{*}{Dyngen Cycle}
& W-4-point (Ours)
  & $2.99 \pm 0.14$
  & $\mathbf{0.04 \pm 0.01}$
  & $0.40 \pm 0.01$
  & $\mathbf{0.07 \pm 0.04}$
  & $\mathbf{0.42 \pm 0.31}$ \\
& F \& S (cubic)
  & $\mathbf{0.03 \pm 0.01}$
  & $\mathbf{0.04 \pm 0.01}$
  & $\mathbf{0.39 \pm 0.01}$
  & $0.08 \pm 0.04$
  & $\mathbf{0.42 \pm 0.31}$ \\
\midrule
\multirow{2}{*}{Uniform Gaussian Cycle}
& W-4-point (Ours)
  & $8.35 \pm 0.75$
  & $\mathbf{3.56 \pm 0.06}$
  & $\mathbf{2.28 \pm 0.02}$
  & $\mathbf{3.19 \pm 2.13}$
  & $\mathbf{1.89 \pm 1.06}$ \\
& F \& S (cubic)
  & $\mathbf{0.07 \pm 0.01}$
  & $4.14 \pm 0.07$
  & $2.52 \pm 0.02$
  & $3.56 \pm 2.52$
  & $2.00 \pm 1.19$ \\
\bottomrule
\end{tabular}
\caption{Quantitative comparison of interpolatory methods for trajectory inference. Best scores are in \textbf{bold}. We compare two interpolatory schemes: F \& S (cubic) splines and our Wasserstein four-point scheme. Both ensure exact interpolation through the observed point clouds.}
\label{tab:interpolation_experiments}
\end{table*}

\end{document}